%% file: neurips_2026.tex
\documentclass{article}

 \usepackage[preprint]{neurips_2026}

\usepackage[utf8]{inputenc} 
\usepackage[T1]{fontenc}    
\usepackage[hidelinks]{hyperref}       
\usepackage{url}            
\usepackage{booktabs}       
\usepackage{amsfonts}       
\usepackage{nicefrac}       
\usepackage{microtype}      
\usepackage{xcolor}         

\usepackage{amsmath}
\usepackage{graphicx}
\usepackage{wrapfig}
\newcommand{\mintinline}[1]{\lstinline[basicstyle=\ttfamily\color{codeblack}]}
\usepackage{caption}
\usepackage{subcaption}
\usepackage{makecell}
\usepackage{listings}
\usepackage{pifont}
\newcommand{\cmark}{\textcolor{green}{\ding{51}}} 
\newcommand{\xmark}{\textcolor{red}{\ding{55}}}   

\definecolor{codegreen}{rgb}{0.0,0.5,0.2}       
\definecolor{codeyellow}{rgb}{0.75,0.55,0.0}    
\definecolor{codeblue}{rgb}{0.0,0.2,0.7}        
\definecolor{codegray}{rgb}{0.5,0.5,0.5}        
\definecolor{codepurple}{rgb}{0.58,0,0.82}      
\definecolor{codeblack}{rgb}{0.1,0.1,0.1}       
\definecolor{backcolour}{rgb}{0.95,0.95,0.92}

\lstdefinelanguage{PythonCustom}{
    language=Python,
    morekeywords={self},
    keywordstyle=\color{codeblue}\bfseries,
    emph={
        numpy,pandas,matplotlib,seaborn,torch,tensorflow,
        sklearn,scipy,os,sys,json,re,math,jax,bluffjax,agent
    },
    emphstyle=\color{codegreen},
    emph={[2]
    print,len,range,input,type,int,float,str,list,dict,set,
        tuple,open,sum,min,max,abs
    },
    emphstyle={[2]\color{codeyellow}},
}

\lstdefinestyle{mystyle}{
    basicstyle=\ttfamily\footnotesize\color{codeblack},
    commentstyle=\color{codegreen},
    stringstyle=\color{codepurple},
    numberstyle=\tiny\color{codegray},
    keywordstyle=\color{codeblue}\bfseries,
    backgroundcolor=\color{backcolour},   
    breaklines=true,
    breakatwhitespace=false,
    keepspaces=true,
    numbers=left,
    numbersep=10pt,
    showspaces=false,
    showstringspaces=false,
    showtabs=false,
    tabsize=4,
    frame=none
}

\title{BluffJAX: Adversarial Imperfect Information Games in JAX}

\author{%
  Aryaman Reddi$^{1,2}$\thanks{Correspondence to \texttt{<aryaman.reddi@tu-darmstadt.de>}}, Jan Peters$^{1,2,3}$, Carlo D'Eramo$^{4}$ \\
  $^1$Department of Computer Science, TU Darmstadt, Germany \\
  $^2$Hessian Center for Artificial Intelligence (Hessian.ai), Germany \\
  $^3$German Research Center for AI (DFKI), Systems AI for Robot Learning, Germany \\
  $^4$Center for Artificial Intelligence and Data Science, University of Würzburg, Germany \\
}

\begin{document}
\maketitle
\input{01_sections/00_abstract}

\input{01_sections/01_intro}
\input{01_sections/02_related}
\input{01_sections/03_bluffjax}
\input{01_sections/04_benchmarking}
\input{01_sections/05_results}
\input{01_sections/06_conclusion}
\bibliography{references}
\bibliographystyle{plainnat}
\newpage
\input{01_sections/091_app_games}
\input{01_sections/092_app_design}
\input{01_sections/093_app_baselines}
\input{01_sections/094_app_head2head}
\newpage
\input{checklist.tex}
\end{document}

%% file: 01_sections/00_abstract.tex
We introduce BluffJAX: an open-source suite of adversarial imperfect information games in JAX. We provide canonical implementations of games designed for high simulation throughputs and parallelization on GPU accelerators. Our suite consists of well-studied benchmarks such as Texas Hold'Em Poker and Kuhn Poker, as well as games that have not been previously studied in reinforcement learning research, such as Bluff, Stud Poker, and Kemps. We hope that implementing a variety of game mechanics and difficulties will introduce new challenges and foster novel research directions in game-theoretic methods for RL. We benchmark the throughput performance and memory usage of our environments in single and multi-GPU settings, demonstrating scaling of up to hundreds of millions of samples per second, and motivating the usage of BluffJAX over related GPU and CPU-based libraries. We benchmark reinforcement learning, tree search, and game-solving algorithms in JAX in order to provide users with baseline results and facilitate future comparisons.

%% file: 01_sections/01_intro.tex
\newcommand{\herowidth}{0.18}
\begin{figure}[htbp]
\centering
\begin{subfigure}[t]{\herowidth\textwidth}
    \centering
    \includegraphics[width=\textwidth]{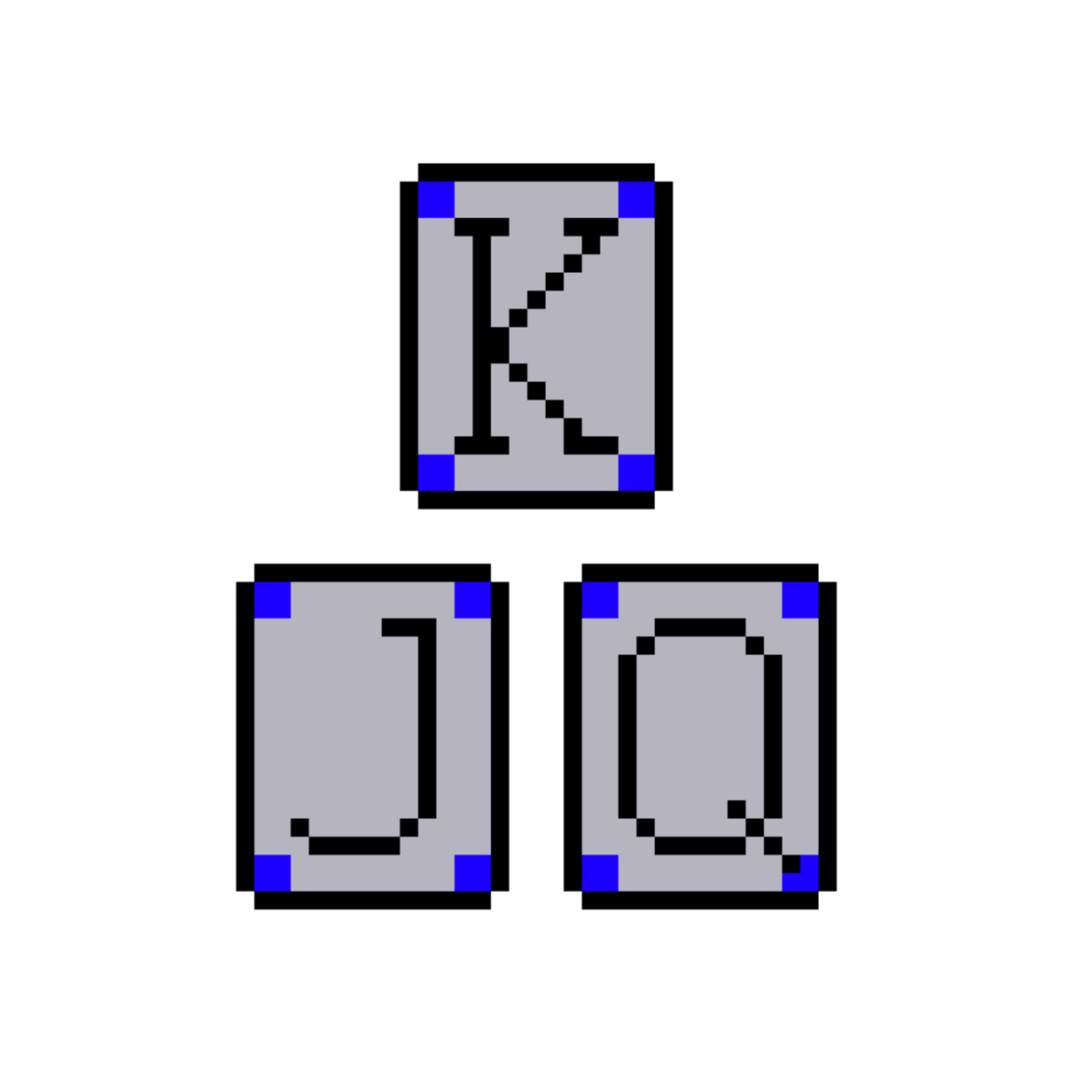}
    \caption{{ Kuhn Poker}}
\end{subfigure}
\begin{subfigure}[t]{\herowidth\textwidth}
    \centering
    \includegraphics[width=\textwidth]{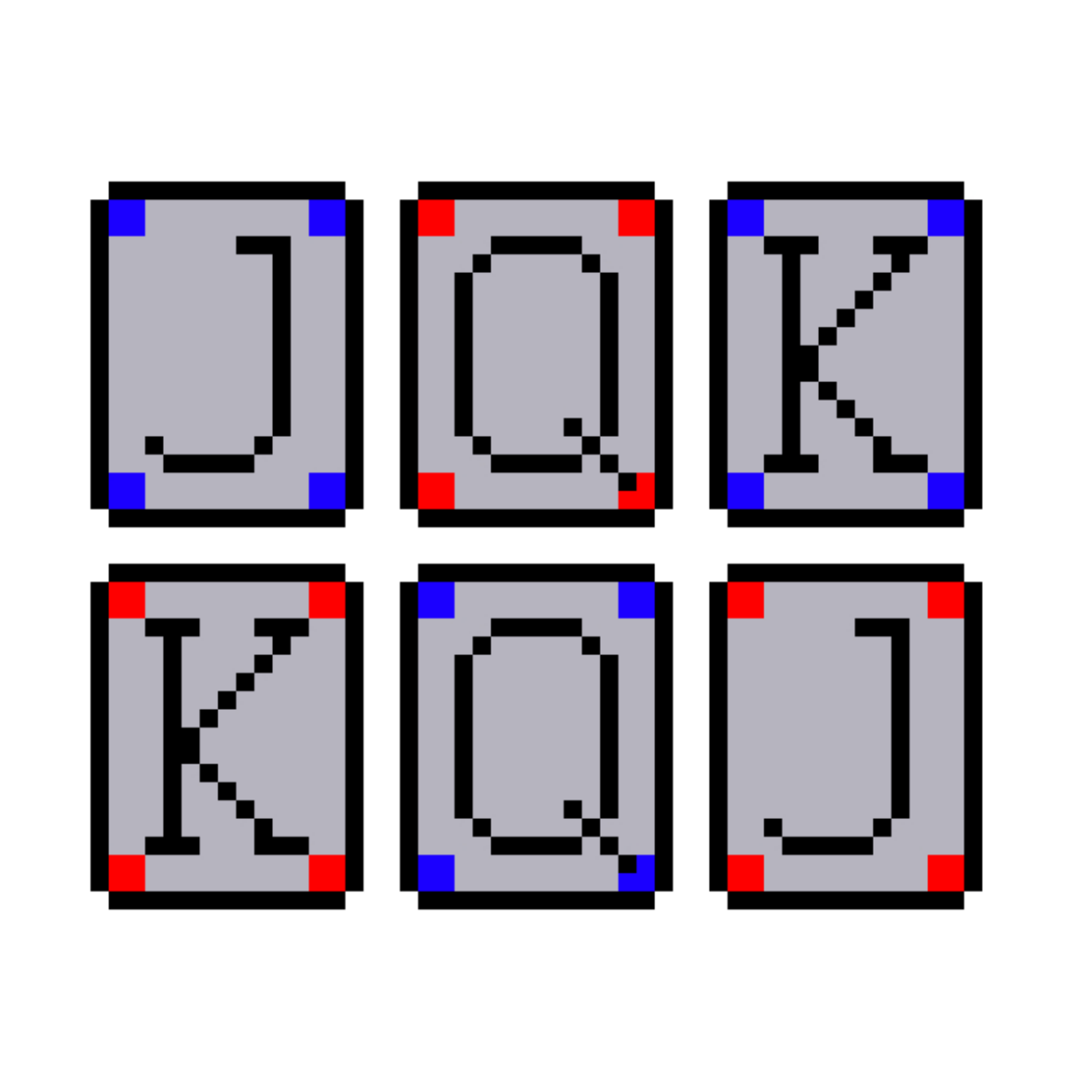}
    \caption{Leduc Poker}
\end{subfigure}
\begin{subfigure}[t]{\herowidth\textwidth}
    \centering
    \includegraphics[width=\textwidth]{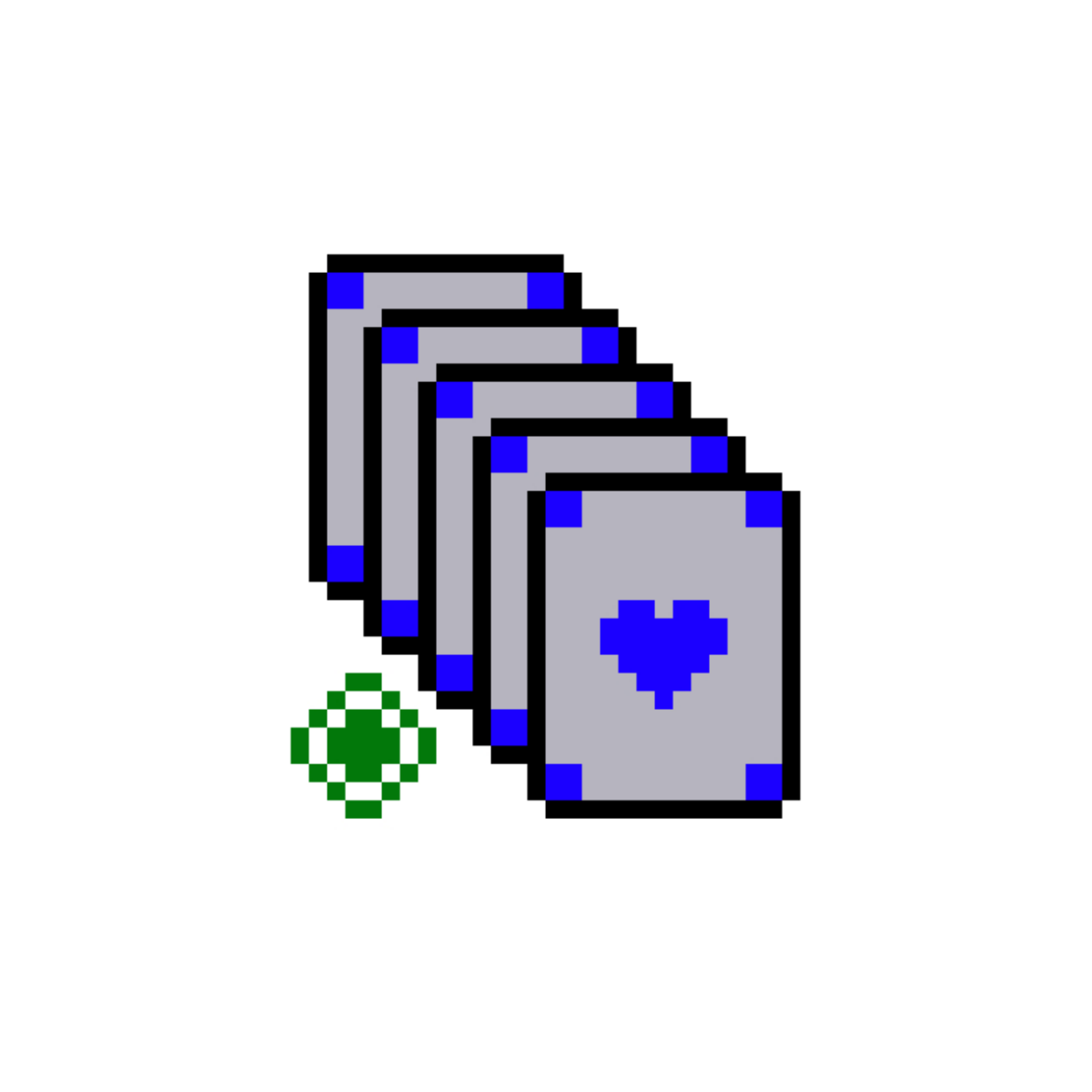}
    \caption{{\centering Texas Hold'Em Limit Poker}}
\end{subfigure}
\begin{subfigure}[t]{\herowidth\textwidth}
    \centering
    \includegraphics[width=\textwidth]{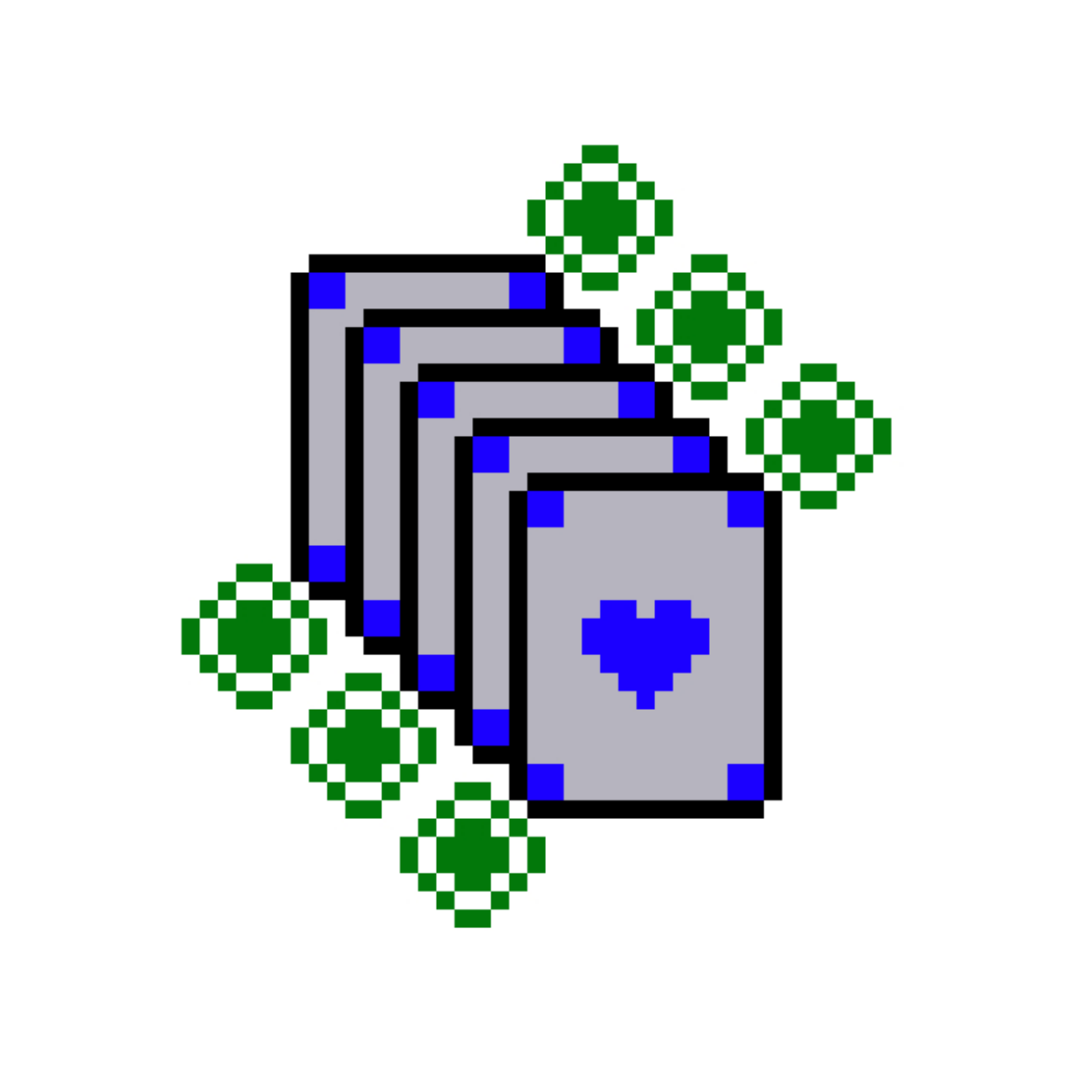}
    \caption{{\centering Texas Hold'Em No-Limit Poker}}
\end{subfigure}
\begin{subfigure}[t]{\herowidth\textwidth}
    \centering
    \includegraphics[width=\textwidth]{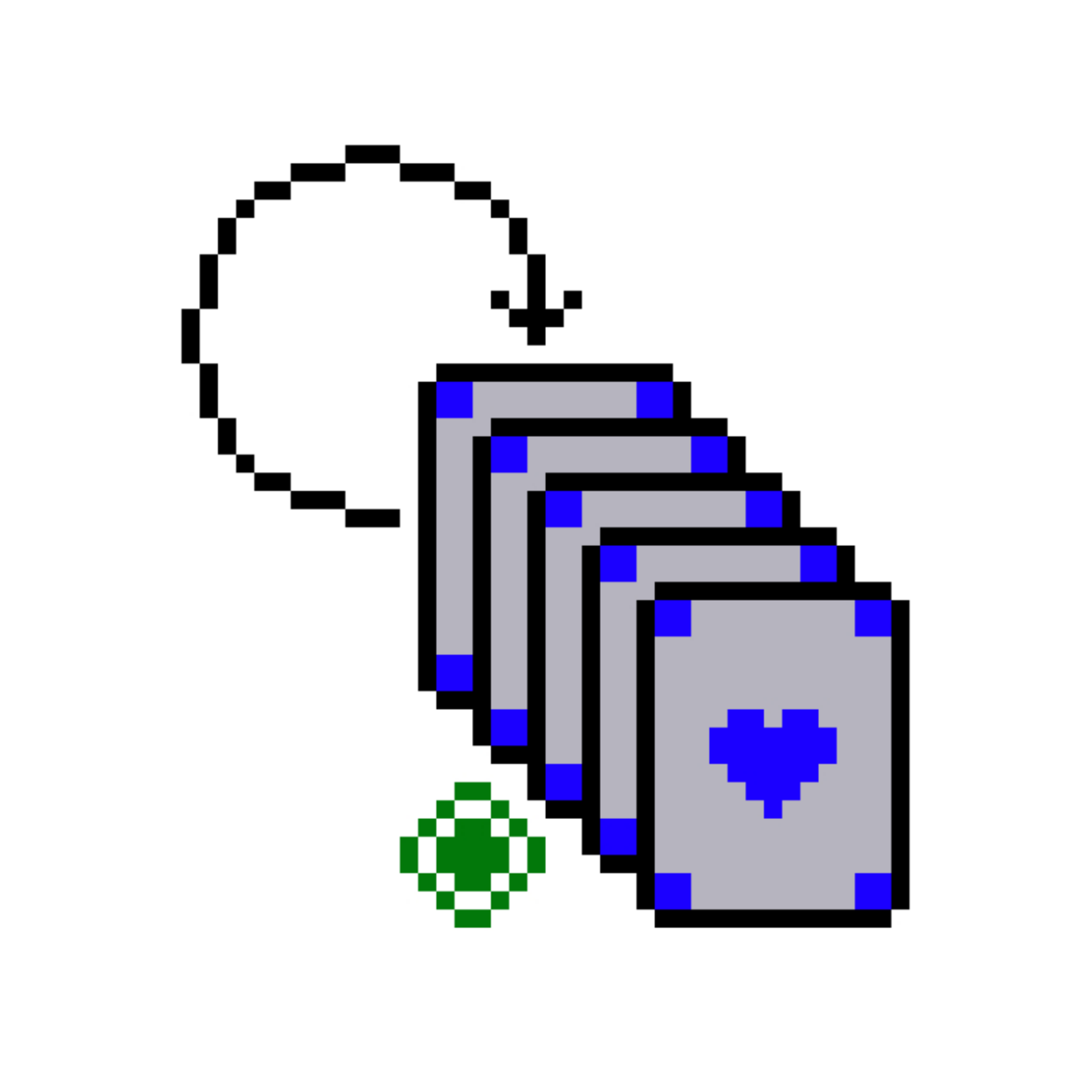}
    \caption{Five Card Draw}
\end{subfigure}

\begin{subfigure}[t]{\herowidth\textwidth}
    \centering
    \includegraphics[width=\textwidth]{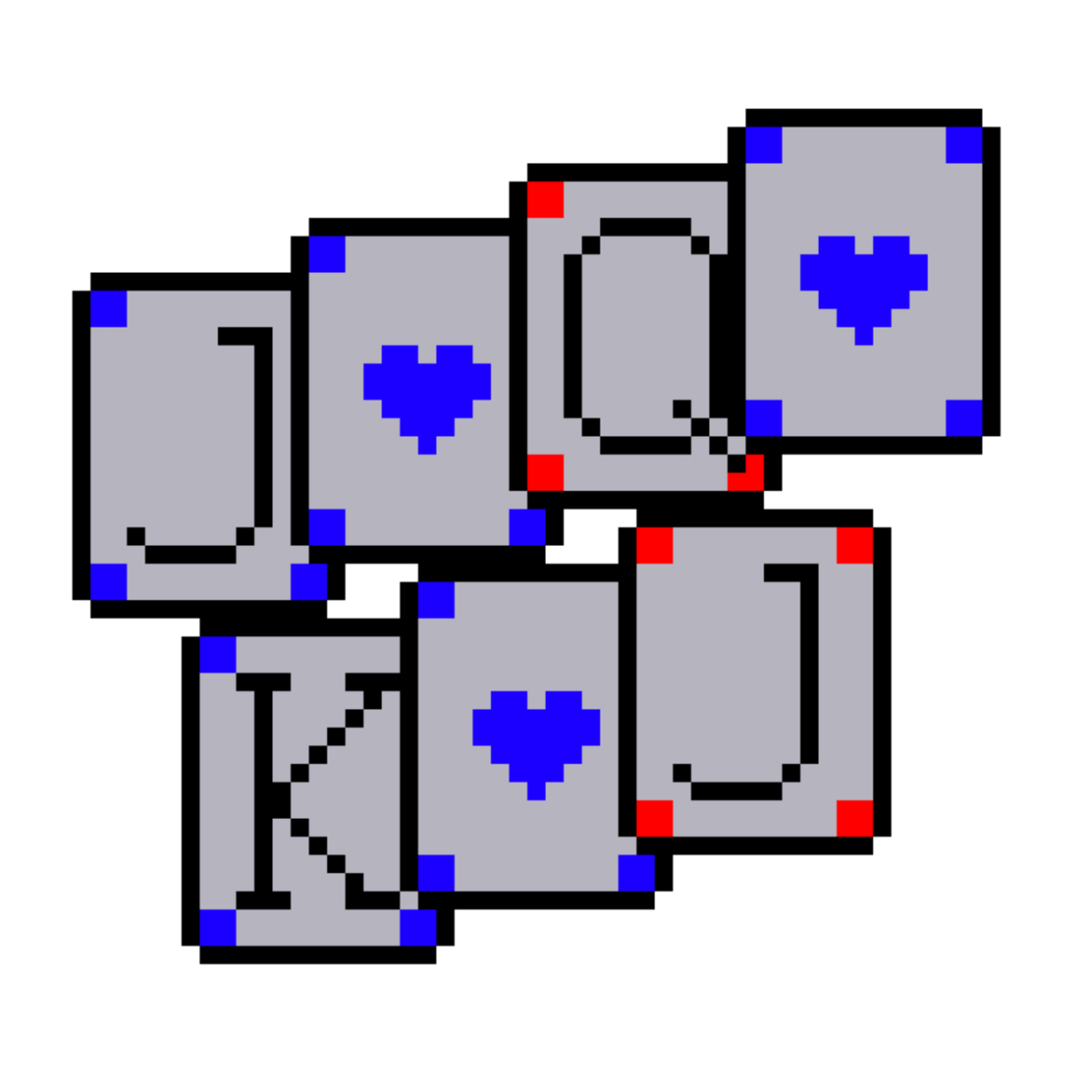}
    \caption{Seven Card Stud}
\end{subfigure}
\begin{subfigure}[t]{\herowidth\textwidth}
    \centering
    \includegraphics[width=\textwidth]{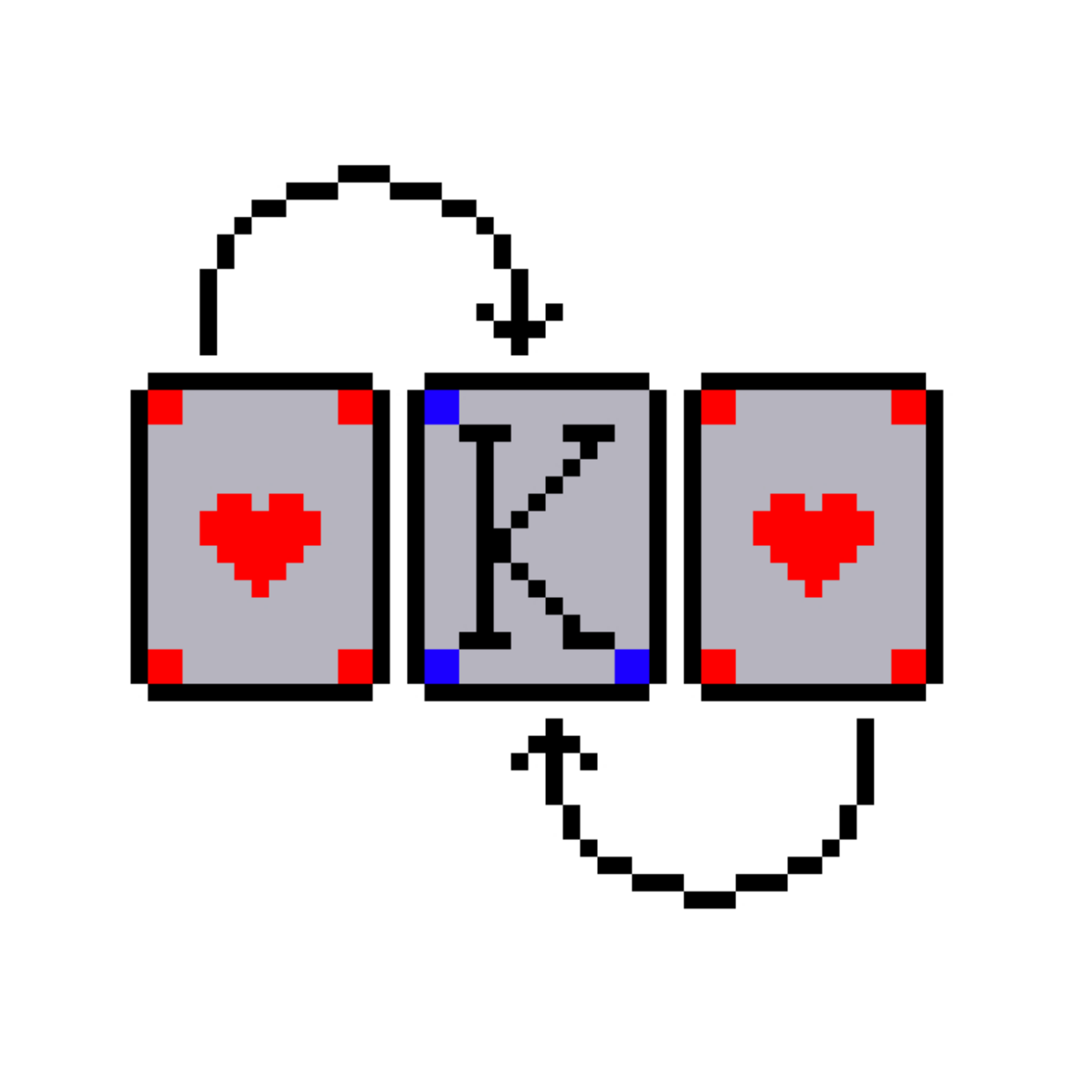}
    \caption{Goofspiel}
\end{subfigure}
\begin{subfigure}[t]{\herowidth\textwidth}
    \centering
    \includegraphics[width=\textwidth]{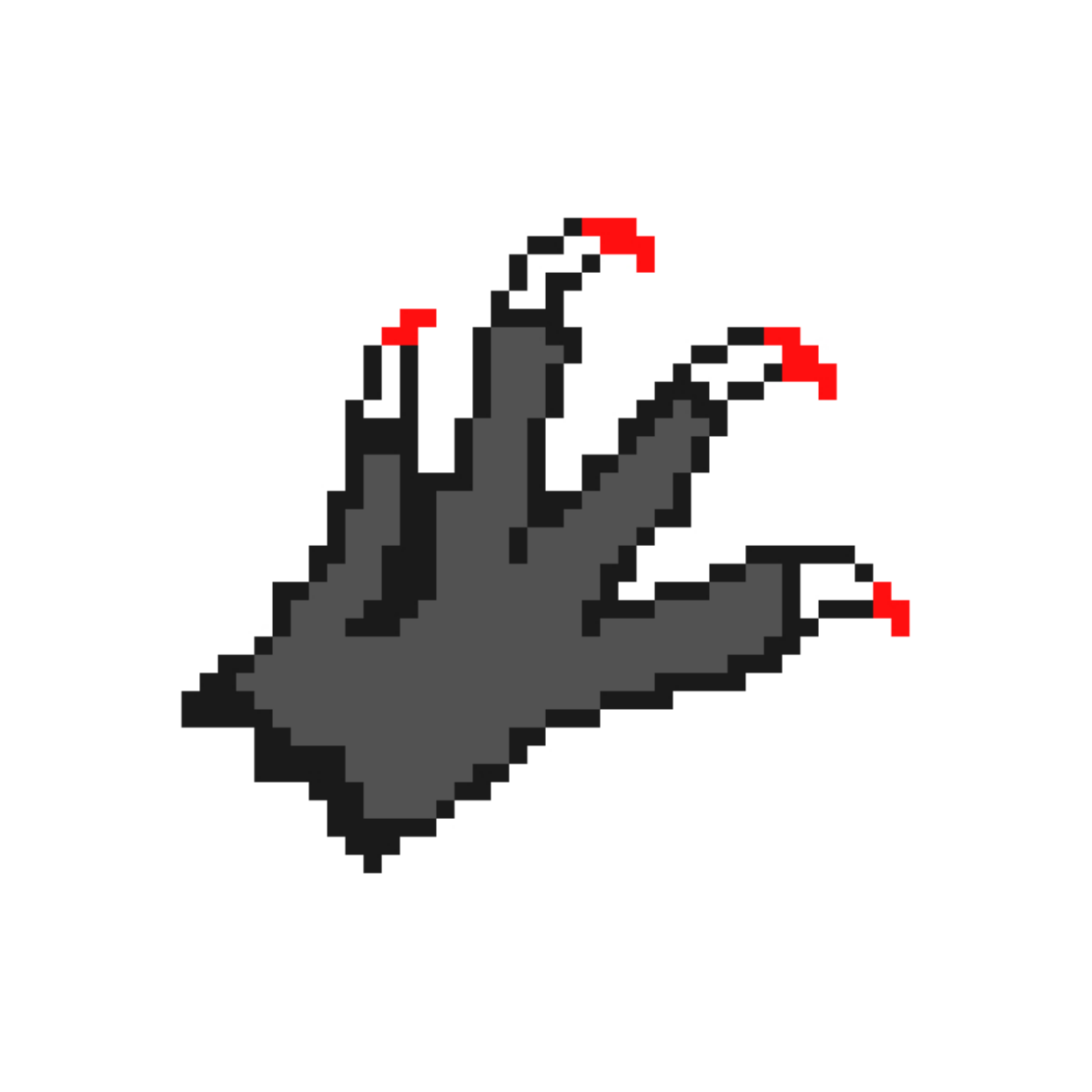}
    \caption{Werewolf}
\end{subfigure}
\begin{subfigure}[t]{\herowidth\textwidth}
    \centering
    \includegraphics[width=\textwidth]{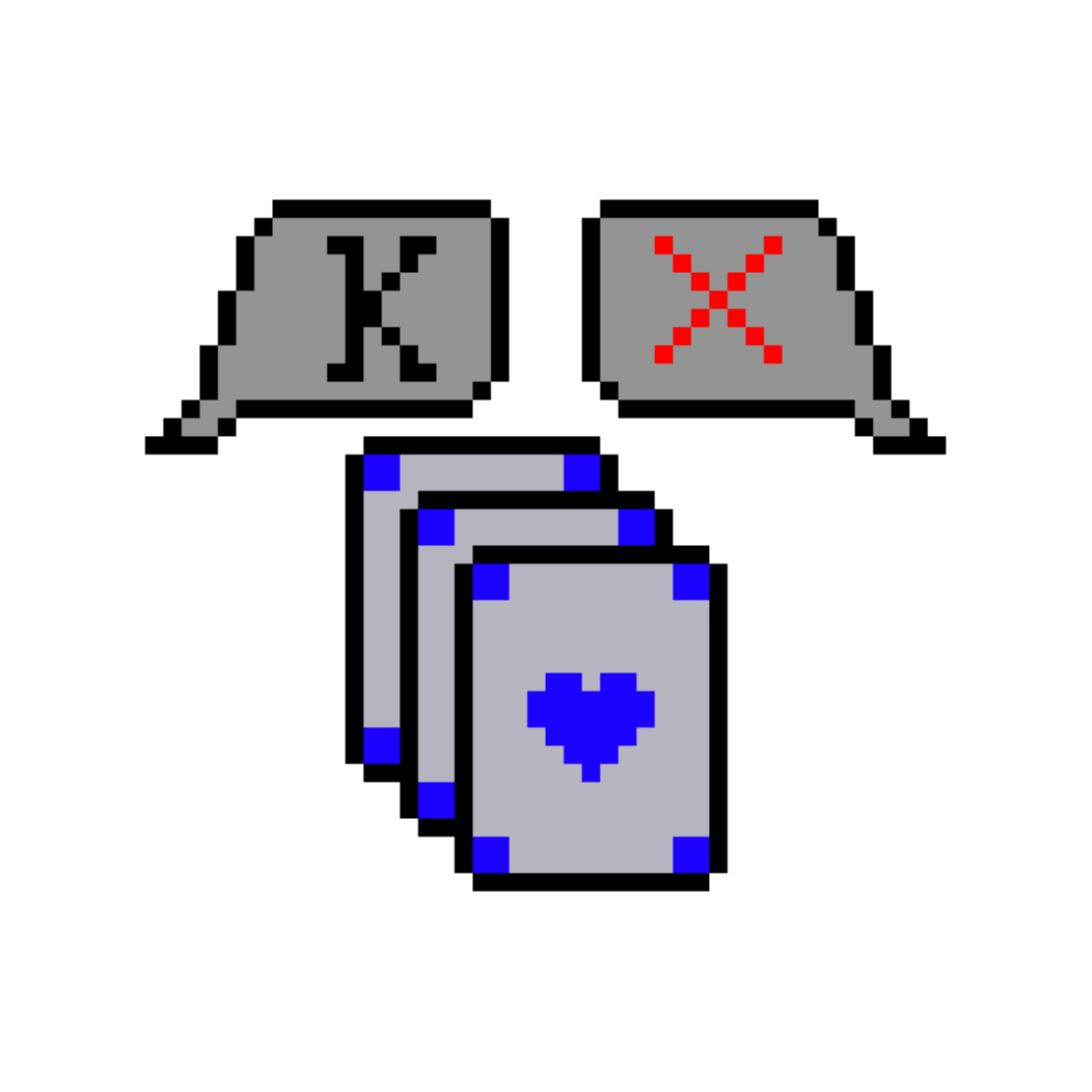}
    \caption{Bluff}
\end{subfigure}
\begin{subfigure}[t]{\herowidth\textwidth}
    \centering
    \includegraphics[width=\textwidth]{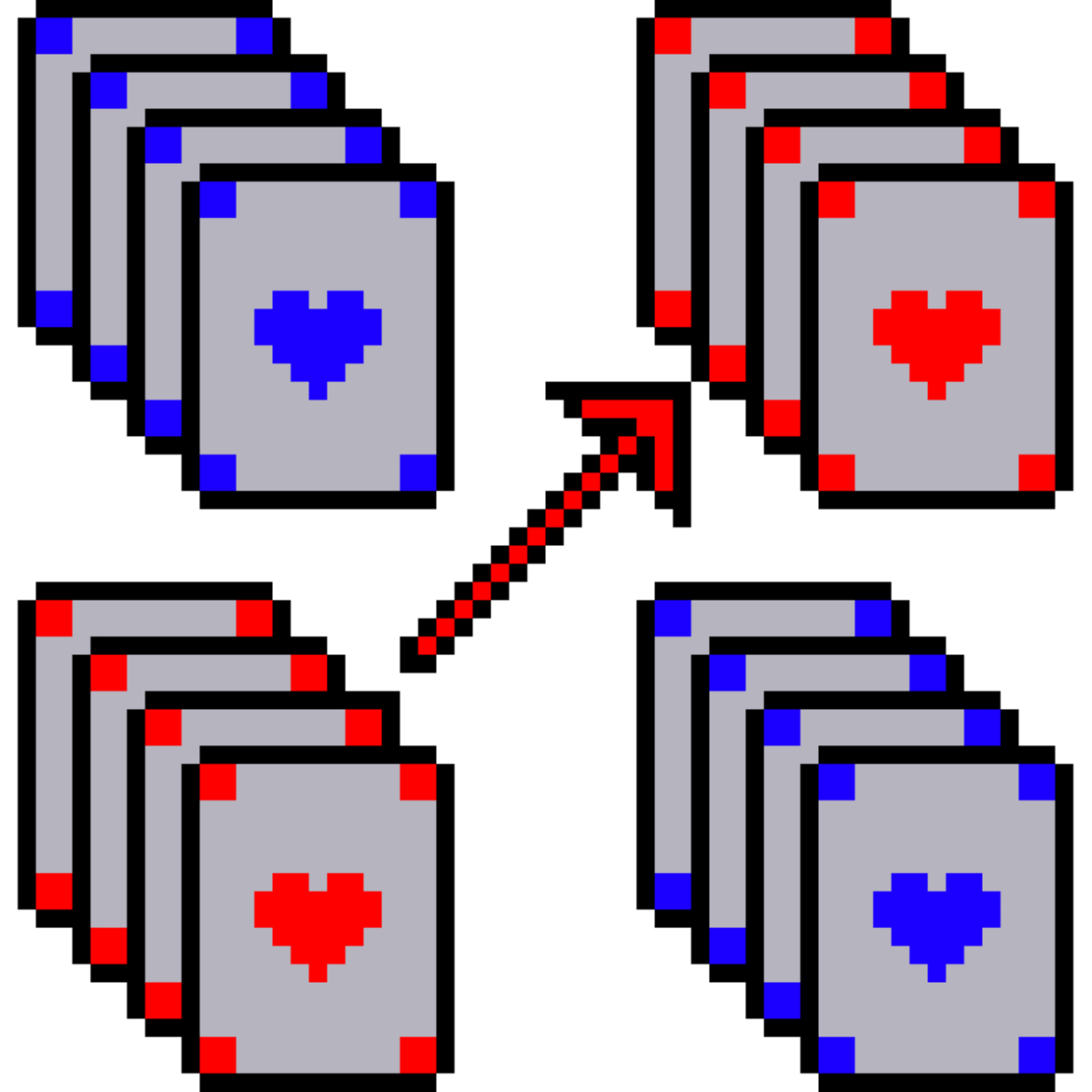}
    \caption{Kemps}
\end{subfigure}

\caption{Games in BluffJAX.}
\label{fig:all_games}
\end{figure}

\section{Introduction}

The development of algorithms that can achieve superhuman performance in games with intractably large state-action spaces has been a driving force in Artificial Intelligence (AI) research for decades~\citep{mnih2015human,silver2016mastering,silver2017mastering}. Imperfect information games present unique challenges: in addition to large state and action spaces, inaccessibility to information and opponent strategy severely hampers standard learning methods like fictitious self-play, value estimation, and tree search - which can be seen in games such as Hold'Em Poker~\citep{brown2020combining}, Diplomacy~\citep{meta2022human}, and Dota 2~\citep{berner2019dota}.

One bottleneck for scaling large-scale RL systems in such domains is the availability of data. High-speed simulators that can leverage hardware accelerators and collect diverse data from parallelized environments are a crucial ingredient in modern RL research. In particular, JAX~\citep{frostig2019compiling} has gained popularity among machine learning researchers in recent years by utilizing the XLA compiler infrastructure for JIT (just-in-time) compilation and automatic differentiation of Python functions. 

\begin{wrapfigure}{r}{0.5\textwidth}
    \centering
    \includegraphics[width=0.5\textwidth]{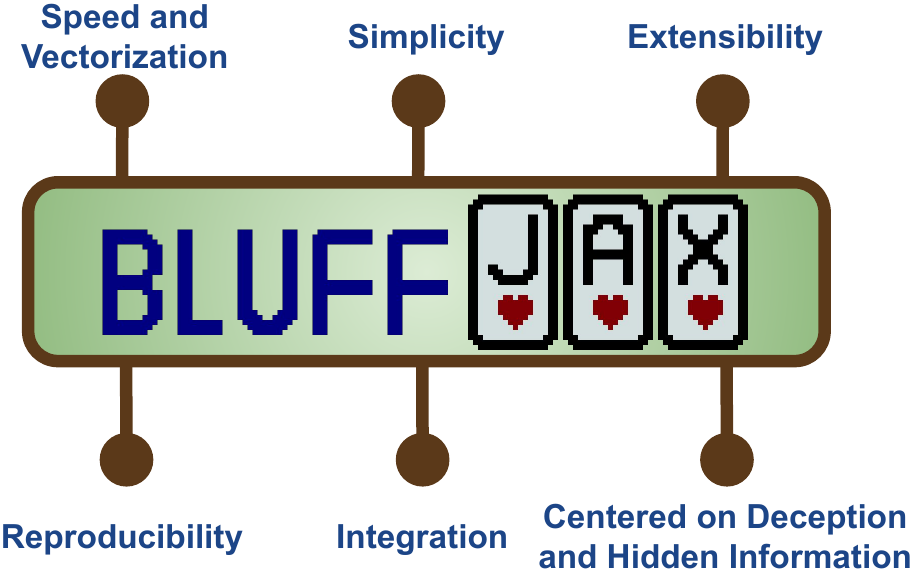}
    \caption{Our philosophy and design criteria.}
    \label{fig:hero}
\end{wrapfigure}

In this work, we introduce \textbf{BluffJAX}, which is (to the best of our knowledge) the first open-source hardware-accelerated JAX suite focused primarily on imperfect information games. Our benchmark include well-studied games such as Texas Hold'Em Poker variants and Goofspiel, while also introducing games that have not been studied in RL research thus far, such as Bluff, Kemps, and multi-round Werewolf. We hope that providing high-speed implementations of well-known games will help alleviate the reproducibility crisis in RL research, as well as introduce new challenges to the RL community that will foster novel research directions in game-theoretic RL and search methods. All code is open-source and available at \url{https://github.com/bluffjax/bluffjax}.

To validate the functionality and scalability of our suite, we benchmark the speed and memory usage of all 10 implemented games across varying numbers of parallel environments in single-GPU and multi-GPU settings. We also analyze the scaling performance of BluffJAX against related game libraries. We provide algorithm baseline results for some well-studied games to motivate our claim that the implementations use canonical game mechanics and are compatible with existing algorithms. Additionally, we include some pre-trained baseline models that we hope will provide users of BluffJAX with a predefined standard by which they can evaluate their own experiments. We implement fully JIT-compatible algorithms for certain games, including Counterfactual Regret Minimization~\citep{zinkevich2007regret} (CFR) and exploitability metrics for Kuhn Poker and Leduc Poker. In large games, it is typically intractable to expand the entire game tree in order to perform CFR; thus, we also implement JIT-compatible deep RL and tree search algorithms.

Figure~\ref{fig:hero} outlines out philosophy and design criteria for BluffJAX. The contributions of our library are summarized as follows:
\begin{itemize}
    \item \textbf{Speed and vectorization:} BluffJAX emphasizes high environment transition throughput rates using efficient game branching logic and leverages JAX-native parallelization, JIT compilation, and on-GPU processing. 
    \item \textbf{Simplicity:} We use a simple API and well-documented code for ease of use.
    \item \textbf{Extensibility:} Our library is structured to enable easy addition of new games and mechanics.
    \item \textbf{Reproducibility:} We provide implementations of well-known game-solving and RL algorithms as well as baseline models to give users a reproducible starting point for their research directions.
    \item \textbf{Integration:} We include wrappers and helper functions to ease the on-boarding of users already familiar with other RL libraries.
    \item \textbf{Centered on Deception and Hidden Information:} To the best of our knowledge, GPU-accelerated implementations of imperfect information games remains a mostly-unfulfilled niche within the ecosystem of RL environment libraries. We aim to fill this gap by providing implementations of both well-known and new games.
\end{itemize}

%% file: 01_sections/02_related.tex
\section{Related Works}
\textbf{Games as AI Benchmarks.} The pursuit of superhuman game-playing abilities in AI systems has been a driving force in AI research for decades~\citep{samuel1959some, tesauro1995temporal, campbell2002deep}, and advances in deep learning have enhanced AI game-playing systems beyond the need for human expert knowledge. ~\citet{mnih2015human} is often attributed with reinvigorating deep RL research by demonstrating superhuman performance in several Atari games using Deep Q-Networks. AlphaGo~\citep{silver2016mastering}, AlphaGo Zero~\citep{silver2017mastering}, Alpha Zero ~\citep{silver2017zero}, and MuZero~\citep{schrittwieser2020mastering} were critical milestones in exhibiting the increasing abilities of deep RL and tree search to reach increasing capabilities in unsolved games like Go, Chess, and Shogi. Systems like OpenAI Five~\citep{berner2019dota}, AlphaStar~\citep{vinyals2019grandmaster}, and Cicero~\citep{meta2022human} have demonstrated the abilities of AI systems to excel in complex multi-agent settings.

\textbf{Algorithms for Imperfect Information Games.} Imperfect information games fundamentally differ from perfect information games, since the existence of private information (partial observability) means that optimal performance requires reasoning over the strategies of other players. As games grow large, tabular methods such as Counterfactual Regret Minimization (CFR)~\citep{zinkevich2007regret} become unfeasible. Various methods have attempted to augment standard RL and tree search algorithms with notions of belief state probability estimation. Deep CFR~\citep{brown2019deep} replaces tabular CFR with Monte Carlo CFR (MCCFR)~\citep{burch2012efficient} for subgame sampling and neural networks to estimate instantaneous regrets. Libratus~\citep{brown2017libratus} and Pluribus~\citep{brown2019superhuman} utilize MCCFR to achieve expert-level performance in 2-player and 6-player Texas Hold'Em Poker respectively. Recursive Belief-based Learning (ReBeL)~\citep{brown2020combining} extended this notion further by leaving state abstraction behind and combining subgame deep CFR with value learning over public belief states. Neural Fictitious Self-Play (NSFP)~\citep{heinrich2016deep} uses best-response policy training with supervised average-policy mimicry from experience replay in order to train RL agents in a 2-player self-play setting, achieving decent performance in model poker games without state abstraction or tree search.

\textbf{The Benefits of JAX.} Over the past few years, adoption of JAX-based training in deep RL research has led to a greater necessity for JAX-native benchmarks. One major benefit of JAX benchmarks is that environment logic can be run on the GPU directly, mitigating the overhead cost of transferring data between the CPU and GPU. The Python JAX function \mintinline{Python}{jax.vmap} allows for efficient parallelization of environments, which has been shown to effectively scale learning~\citep{gallici2024simplifying}. Finally, \mintinline{Python}{jax.jit} allows for just-in-time compilation of JAX-compatible Python functions using the XLA compiler, leading to demonstrable speedups over PyTorch-based neural network training and array logic.

\textbf{Related Benchmarks.} Suites such as Gymnax~\citep{gymnax} and Jumanji~\citep{bonnet_jumanji_2024} implement well-known single-player environments into JAX and have provided researchers with familiar RL environments such as classic control games and MinAtar. Brax~\citep{freeman2021brax} implements differentiable rigid-body simulation in JAX for physics-based control, enabling faster training in robotic domains. JAXMarl~\citep{rutherford_jaxmarl_2024} implements well-studied multi-agent environments such as StarCraft II, Multi-Particle Environments, and Hanabi. MEAL~\citep{tomilin_meal_2025} presents a benchmark tailored to continual MARL by procedurally generating tasks based on the Overcooked benchmark. PGX~\citep{koyamada_pgx_2023} implements adversarial games with (mostly) perfect information in JAX, such as Chess, Go, and Hex. RLCard~\citep{zha_rlcard_2020} and PettingZoo~\citep{terry2021pettingzoo} are CPU-based suites that contain some overlapping environments with BluffJAX (e.g. Texas Hold'Em Poker). Finally, OpenSpiel~\citep{lanctot2019openspiel} is worth noting as it provides canonical implementations of several games in C++ and has been used across many works\footnote{It should be noted that some JAX code examples (e.g. CFR, DQN) are provided in the OpenSpiel library; however, the games themselves are CPU-based.}. In this work, we address an open niche in the benchmark ecosystem, since \textbf{BluffJAX specializes on JAX implementations of adversarial games with imperfect information, particularly those involving deception and signaling}. Table~\ref{tab:benchmarks} summarizes some notable attributes of BluffJAX when compared to other well-known MARL benchmarks. Note that we describe BluffJAX as \textit{mostly} adversarial because of the inclusion of team games that require collaboration within each team. Table~\ref{tab:benchmarks} comapres BluffJAX to other relevant suites, including those mentioned above as well as~\citet{ruhdorfer2024overcooked, agapiou2022melting,kurach2020google}.

\begin{table}[htbp]
    \caption{Comparison between BluffJAX and other Multi-Agent Benchmarks.}
    \begin{center}
        \begin{tabular}{lccllc}
        \hline
            Benchmark  & \makecell{GPU-\\accelerated}  &\makecell{Variable \\
            no. agents} & \makecell{Adversarial or\\cooperative} & \makecell{Perfect or \\ Imperfect Info.}  
            \\ \hline
            PGX & \cmark & \xmark & Adversarial (mostly) & Perfect (mostly) \\
            JaxMARL& \cmark & \cmark & Both & Both  \\
            Overcooked GC & \cmark & \xmark & Cooperative & Perfect \\
            MEAL & \cmark & \cmark & Cooperative & Both &  \\
            VMAS  & \xmark & \cmark & Both & Imperfect \\
            OpenSpiel & \xmark & \cmark & Both & Both  \\
            RLCard & \xmark & \xmark & Adversarial & Imperfect  \\
            PettingZoo & \xmark & \cmark & Both & Both  \\
            Melting Pot 2 & \xmark & \cmark & Both & Both  \\
            Google Football & \xmark & \cmark & Both & Both  \\
            \hline
            BluffJAX & \cmark & \cmark & Adversarial (mostly) & Imperfect \\
            \hline
        \end{tabular}
    \end{center}
    \label{tab:benchmarks}
\end{table}

%% file: 01_sections/03_bluffjax.tex
\section{BluffJAX}
In this section, we discuss our choices of environments, API, and design elements of BluffJAX. Full details of the games, including mechanics, observation/action spaces, and rewards are available in Appendix~\ref{supp:games}. A thorough discussion of the API and particular design choices is available in Appendix~\ref{supp:design}. 

\subsection{Environments}
We aim for a mixture of toy environments, well-studied games, and new games in BluffJAX. A summary of the 10 implemented games is available in Table~\ref{tab:games}, where we note the number of agents, observation size, number of actions, number of JAX branches, and the performance metric for each game. Note that the number of JAX branches refers to the number of parallel executions at each timestep in game logic, which is not necessarily equal to the branching factor of the game.

\textbf{Kuhn Poker}~\citep{kuhn1953extensive} and \textbf{Leduc Poker} are well-studied and theoretically solved poker variants, in that their exact exploitability and equilibrium strategies are tractable. In this paper, we benchmark some common RL and game-theoretic algorithms on Kuhn and Leduc Poker in order to validate our implementations and motivate our choices for baselines in more complex environments.

\textbf{Texas Hold'Em Limit Poker} and \textbf{Texas Hold'Em No-Limit Poker} are commonly-studied poker variants which present sufficient challenges to existing game-solving algorithms. They are typically implemented with certain abstractions for simplification; e.g. fixed increment sizes and simple bankroll management. 

We also introduce two lesser-known poker variants that include mechanics that we believe pose interesting challenges for current RL algorithms. In \textbf{5 Card Draw Poker}, players build a hand between betting rounds by exchanging cards with the house, which moves the signaling focus (which is typically on bet sizes and public cards) to the exchanges made by players. The exchange mechanic introduces a new dimension of play, as agents must consider more sources of deception with less information. \textbf{7 Card Stud Poker} is a variant where player hands are revealed across multiple betting rounds, requiring reasoning based on public information and betting history. Stud Poker differs to Texas Hold'Em as a game with longer horizons, higher information density, and more dynamic strategies.

\textbf{Goofspiel} is a 2-player simultaneous-move game which simulates simple auctions; each round consists of players making a fixed bet for a random, visible prize until no prizes or bids remain. Pure strategies in Goofspiel are strictly dominated, requiring agents to model stochastic strategies in order to avoid exploitation.

\textbf{Bluff (AKA Cheat, I Doubt It)} is a card game where each player aims to reduce their hand to 0 by discarding cards in a constrained order - players have the option to lie about which cards they are discarding. Bluff presents a unique challenge among the BluffJAX environments, as agents must evade detection in long episodes in order to succeed.

\textbf{Werewolf} is a team-based social role game that has been studied in the context of reasoning and cooperation among Large Language Models. In BluffJAX, we introduce the first JAX-based implementation of multi-round werewolf with multiple roles. As a social deduction game, werewolf requires agents to cooperate with other members of their respective team in order to win. \textbf{Kemps} is a team game centered around communication where agents play in teams of two and aim to build a hand of a certain rank. To win, they must publicly signal their teammate about their complete hand at the risk of getting caught by opposing teams.

\begin{table}[htbp]
    \caption{Games in BluffJAX. Note than in several games, the observation size depends on n. Note also that Kemps contains an explicit communication channel (default size 2).}
    \begin{center}
        \begin{tabular}{lrrrrl}
        \hline
            \multicolumn{1}{c}{Environment}  &\multicolumn{1}{c}{n}  &\multicolumn{1}{c}{Obs. size} &\multicolumn{1}{c}{\# Actions} &\multicolumn{1}{c}{\# Branches}  &\multicolumn{1}{c}{Metric} 
            \\ \hline
            Kuhn Poker & 2 & 9 & 2 & 1 & exploitability   \\
            Leduc Poker & 2 & 36 & 3 & 3 & exploitability  \\
            THL Poker & 2-10 & 72+n & 4 & 4 & chips/hand  \\
            THNL Poker & 2-10 & 54 & 5 & 5 & chips/hand  \\
            5 Card Draw  & 2-10 & 54 & 37 & 11 & chips/hand  \\
            7 Card Stud  & 2-10 & 77+208(n-1)+n & 4 & 4 & chips/hand  \\
            Goofspiel & 2 & 39 & 13 & 1 & win rate  \\
            Bluff & $\geq$3 & 263 & 13 & 10 & win rate  \\
            Werewolf & 6 & 43 & 7 & 7 & win rate  \\
            Kemps & 4,6,8.. & 52(n+1) + n(comm) & 172*comm. & 1 & win rate \\
            \hline
        \end{tabular}
    \end{center}
    \label{tab:games}
\end{table}

\subsection{API}
The API supports both parallel and agent-environment-cycle environments (AEC) with generic base classes that can be easily modified in order to fulfill specific game logic. Unlike other suites which use a dual API format~\citep{terry2021pettingzoo}, BluffJAX forgoes agent iterator cycles in favor of handling agent transitions and reset logic internally for ease of use. For ease of use for researchers more familiar with other libraries, wrappers for the APIs of JaxMARL and PettingZoo are available. Since most research in imperfect information games is done in self-play, we find that the most efficient pattern is to keep the agent-environment interface agent-agnostic by default, allowing specific placement of agents if required. Games such as Bluff, Goofspiel, and Kemps are easily customizable and can be run at different difficulty levels by simply changing certain environments parameters.

The following code snippet shows the basic usage of BluffJAX to run 1000 agent actions in 16 parallel environments:

\begin{lstlisting}[language=Python]
import jax
import bluffjax
import agent

env=make("kuhn_poker") # Create an environment
rng=jax.random.key(seed=42)
rng_resets=jax.random.split(rng,16) # 16 parallel environments
state,obs=jax.vmap(env.reset)(rng_resets)

def step(carry,unused):
    state,obs=carry
    action=jax.vmap(agent.get_action)(obs)
    state,obs,rew,absorbing,done,info=jax.vmap(env.step)(state,action)
    return (state,obs),rew

def step_scan(state,obs):
    return jax.lax.scan(step,(state,obs),None,length=1000)
    
(final_state,final_obs),rew=jax.jit(step_scan)(state,obs)
\end{lstlisting}

We use this snippet to explain some features of our API and functional JAX programming for those unfamiliar. The object \mintinline{Python}{rng} is a pseudorandom key generated deterministically by an initial seed. Unlike NumPy, JAX requires explicitly passing pseudorandom keys to functions that require randomness\footnote{https://docs.jax.dev/en/latest/random-numbers.html}. \mintinline{Python}{jax.random.split} deterministically splits a pseudorandom key into multiple keys that may be used by separate random functions. \mintinline{Python}{jax.vmap} is a vectorizing map which maps a function over vectorized arguments - in this case, \mintinline{Python}{env.reset} is vectorized over an array of reset keys, producing vectorized states and obs arrays. \mintinline{Python}{state} contains all the information that defines the current game state, including the current player and public/private information. \mintinline{Python}{obs} is a vector observation for the current player. \mintinline{Python}{env.step} takes the current state and action and generates the next state, the observations for the next player, and a vector of rewards for all agents. The function \mintinline{Python}{step} is converted into a \mintinline{Python}{jax.lax.scan} function with \mintinline{Python}{length=1000} to indicate 1000 sequential steps within each parallel environment. Finally, \mintinline{Python}{step_scan} outputs the final carry and vectorized reward outputs.

%% file: 01_sections/04_benchmarking.tex
\section{Performance Benchmarking}
\label{sec:benchmarking}

\begin{figure}[htbp]
    \centering
        \centering
        \includegraphics[width=\textwidth]{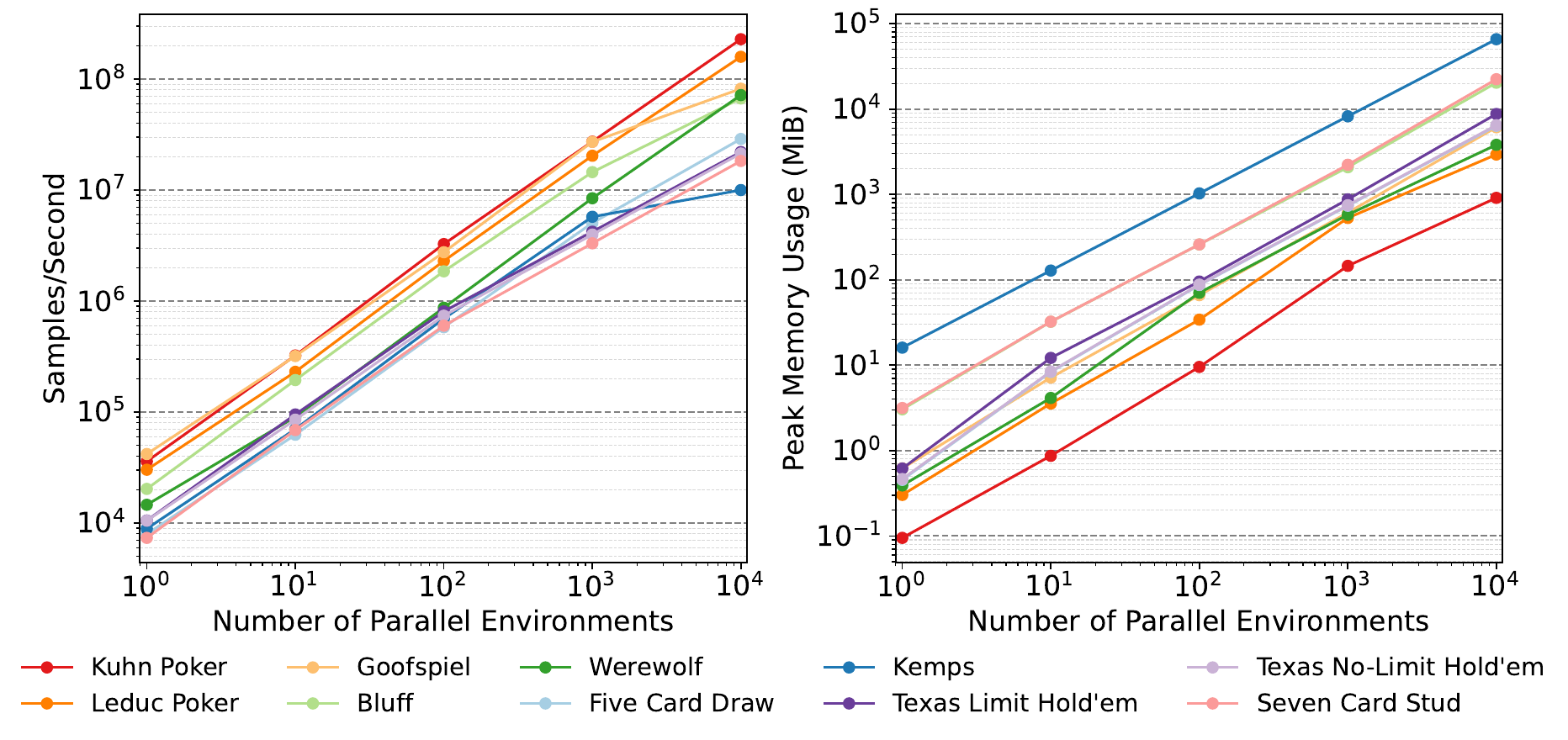}
        \caption{Speed and peak memory usage benchmarking for BluffJAX environments. All results are obtained on an RTX 6000 Ada GPU for 1000 random steps on all parallel environments. Each datapoint is a mean across 10 seeds (standard errors are not visible at this scale).}

    \label{fig:speed_memory}
\end{figure}

\begin{figure}[htbp]
    \centering
        \centering
        \includegraphics[width=\textwidth]{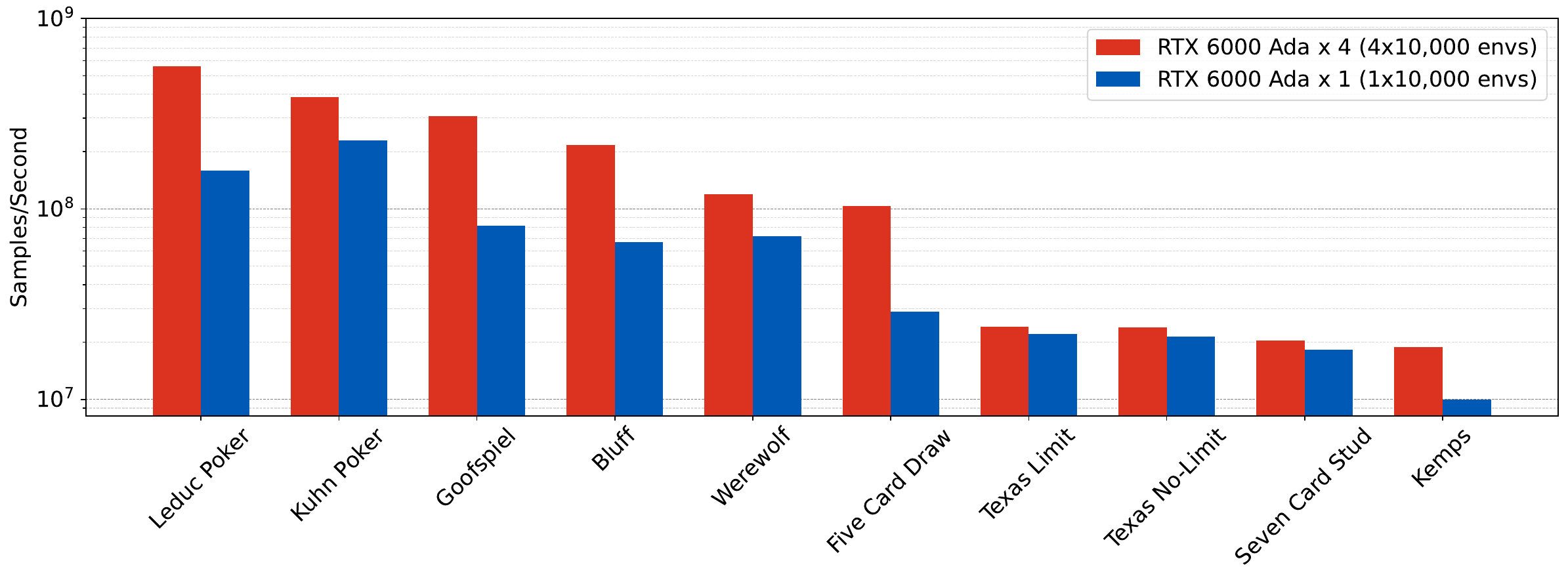}
        \caption{Speed benchmarking for BluffJAX environments on single vs multi-GPU setups. All results are obtained with RTX 6000 Ada GPUs for 1000 random steps on 10,000 parallel environments. Each bar is a mean across 10 seeds (standard errors are not visible at this scale).}
    \label{fig:speed_gpus}
\end{figure}

In this section, we measure the speed and memory requirements of BluffJAX environments. Figure~\ref{fig:speed_memory} shows the simulation throughput and peak memory usage for increasing parallel environments for all BluffJAX games. For almost all games in BluffJAX, throughput scales log-linearly with the number of parallel environments. At 10,000 environments, we see that even the slowest game (Seven Card Stud) still generates on the order on $\sim$2e7 samples/s, evincing the benefits of parallelization in addressing data bottlenecks. Certain environments scale sub-log-linearly in throughput at high parallelization, which appears to be due to GPU memory constraints, as peak memory usage scales log-linearly as well - notably, Kemps demands the highest memory and throughput suffers after 1000 environments. As expected, environments with heavier game logic and larger array operations are subject to higher memory requirements and lower throughput at all levels of parallelization. In Figure~\ref{fig:speed_gpus}, we compare simulation throughput on both single and multi-GPU settings. We see that in the availability of high compute resources, BluffJAX achieves an average of 2.3x higher throughput.

\begin{figure}[htbp]
    \centering
        \centering
        \includegraphics[width=\textwidth]{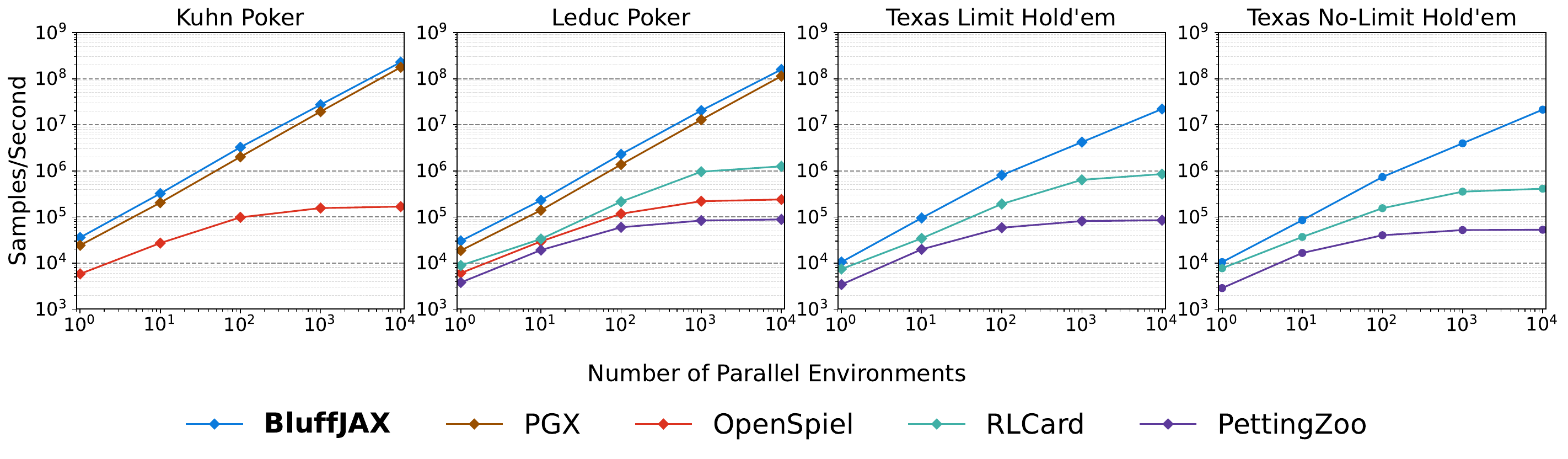}
        \caption{Speed benchmarking comparison between BluffJAX and related libraries on single-GPU and multi-CPU-core setups. We specifically chose the 4 games with the most overlapping implementations in related libraries (although only BluffJAX implements all 4). All GPU results are obtained on an RTX 6000 Ada GPU for 1000 random steps on all parallel environments. All CPU results are obtained on a computational cluster with 64GB of RAM and an AMD Ryzen 9 16-Core processor using the Python multiprocessing library. Each datapoint is a mean across 10 seeds (standard errors are not visible at this scale).}
    \label{fig:speed_other}
\end{figure}

Figure~\ref{fig:speed_other} compares the sample throughput rates of BluffJAX against those of similar GPU-based (PGX) and CPU-based (OpenSpiel, RLCard, PettingZoo) libraries.  The throughputs of BluffJAX are similar to those of PGX - this is likely due to the relative simplicity of the overlapping games (Kuhn/Leduc poker) which leaves little room for further game logic optimizations. Our similarity to PGX motivates our claim that our implementations are canonical and well-optimized, since they match the results of a well-established benchmark also designed for GPU acceleration. Note that our novel contribution relative to PGX is not in the throughput rates for these 2 simple games, but rather in our JAX implementations of non-overlapping games. BluffJAX outperforms all CPU-based libraries with a gap of at least an order of magnitude at 1000 environments and beyond. Notably, efficient vectorization enables GPU-based libraries to continue scaling linearly as parallelization increases, while scaling stagnates on CPUs. Note also that BluffJAX outperforms CPU-based libraries even with no parallelization due to efficient game tree logic and leveraging JIT compilation.

%% file: 01_sections/05_results.tex
\section{Experimental Results}
\label{sec:experiments}
In this section, we validate our implementations and benchmark some RL, tree search, and model-based game solving algorithms on our implemented environments. We first validate the ability of our JAX-based implementation of CFR to find near-optimal strategies for solved games, as well as standard RL, RL augmented with NFSP, and Deep CFR algorithms. We then compare RL+NFSP with ReBeL baselines on Texas Hold'Em Poker, a well-studied medium for game-theoretic algorithms. Finally, we benchmark our remaining environments with RL+NFSP baselines.

\begin{figure}[htbp]
    \centering
    \begin{subfigure}[b]{0.45\textwidth}
        \centering
        \includegraphics[width=\textwidth]{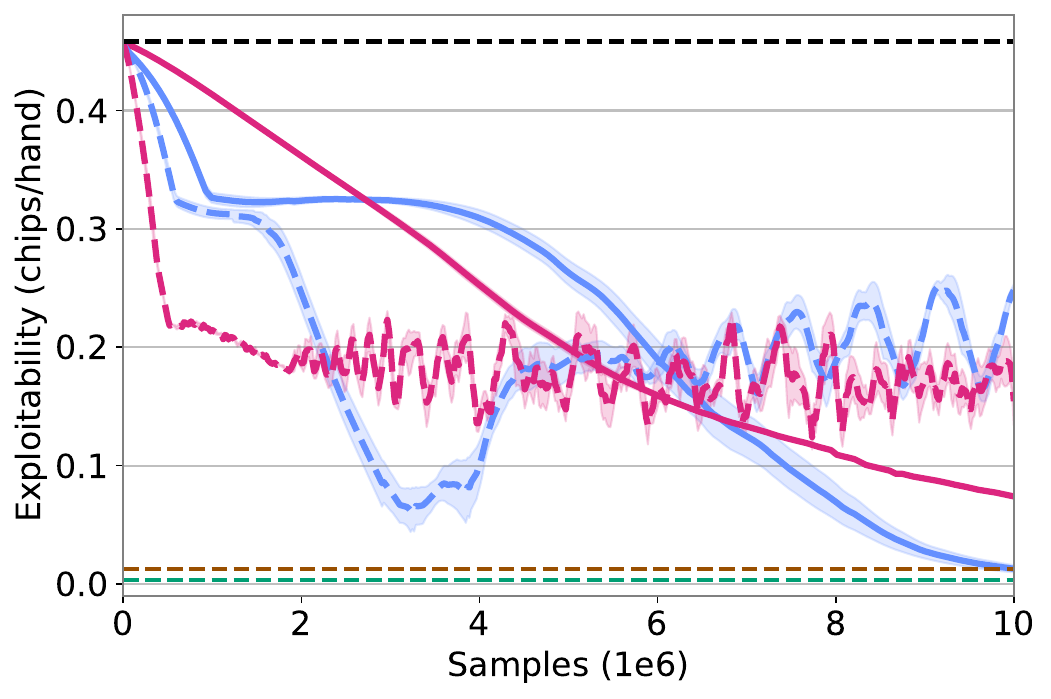}
        \caption{Kuhn Poker}
        \label{fig:kuhn}
    \end{subfigure}
    \begin{subfigure}[b]{0.45\textwidth}
        \centering
        \includegraphics[width=\textwidth]{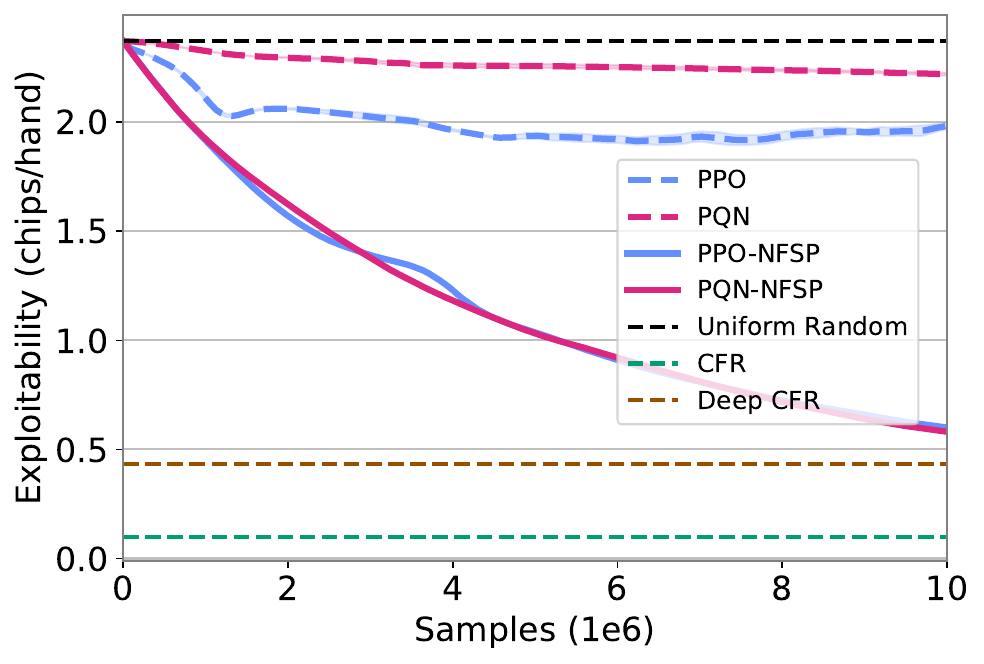}
        \caption{Leduc Poker}
        \label{fig:leduc}
    \end{subfigure}
    \caption{Exploitability of Kuhn and Leduc Poker. Each algorithm is trained in self-play with 1e7 samples and evaluated across 10 seeds.}
    \label{fig:kuhn_leduc}
\end{figure}

Figure~\ref{fig:kuhn_leduc} shows the exploitability of various algorithms on Kuhn and Leduc Poker. The exploitability of a strategy profile is defined as net utility gained by an optimal policy playing against that strategy profile. Lower exploitability is better; an exploitability of $0$ means a strategy profile is a Nash equilibrium~\citep{nash1950equilibrium}. We note the exploitability of a uniform random policy on Kuhn ($0.4583$) and Leduc ($2.3736$), which matches values from \citet{kawamura2017neural} and OpenSpiel. We also include the exploitability obtained by $100$ iterations of CFR. Deep CFR achieves near-convergence on Kuhn Poker with $100$ iterations, but struggles with Leduc Poker. This may be because truncating the relatively shallow game tree of Leduc Poker may exaggerate errors from function approximation.

We benchmark the exploitability of two RL algorithms, Proximal Policy Optimization (PPO)~\citep{schulman2017proximal} and Parallelized Q-Network (PQN)~\citep{gallici2024simplifying}, on these environments. We choose PPO and PQN as they are simple to implement, well-known by the RL research community, and are well-suited to parallelized environment training and GPU acceleration. While PPO and PQN are able to achieve decent results in Kuhn Poker (beating Uniform Random by a fair margin), they tend to exhibit oscillatory behavior. Cycling is a well-studied phenomenon in two-player zero-sum games~\citep{shapley1963some, balduzzi2018mechanics} which is worsened in self-play settings due to bootstrapping with self-produced training data~\citep{heinrich2016deep}.

We also implement PPO-NFSP and PQN-NFSP, which are versions of NFSP with PPO and PQN as the best response learners respectively alongside a supervised anticipatory network. We note that they achieve better results than vanilla PPO and PQN in both environments, but still fall short of optimality in Leduc Poker. 

\begin{figure}[htbp]
    \centering
    \begin{subfigure}[b]{0.4\textwidth}
        \centering
        \includegraphics[width=\textwidth]{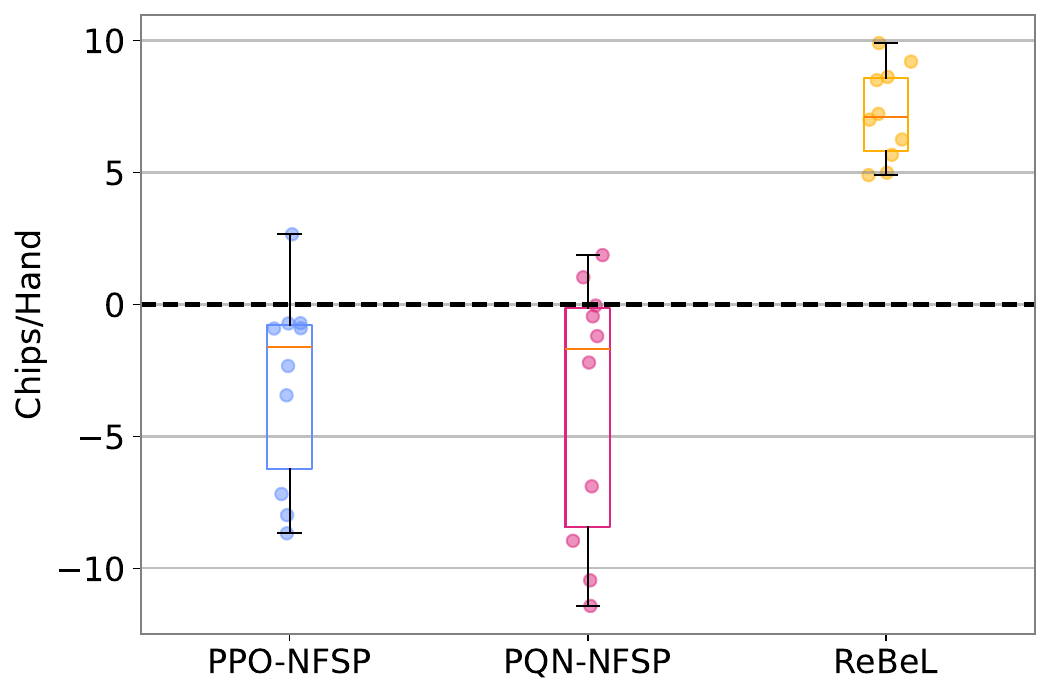}
        \caption{Heads-Up Texas Hold'Em Limit}
        \label{fig:hul}
    \end{subfigure}
    \begin{subfigure}[b]{0.4\textwidth}
        \centering
        \includegraphics[width=\textwidth]{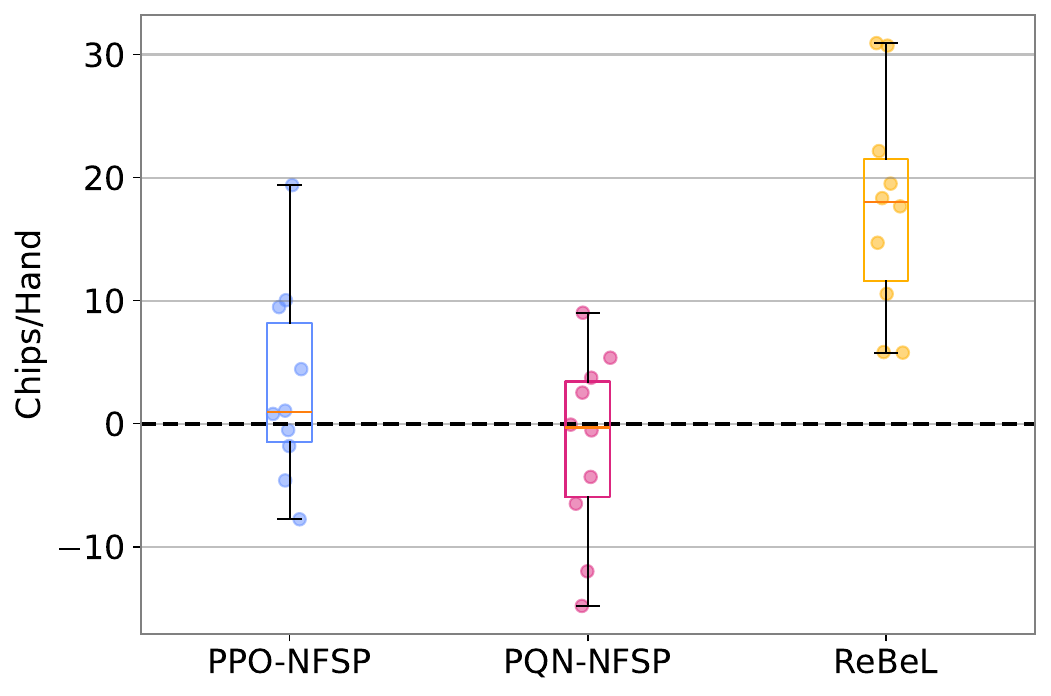}
        \caption{Heads-Up Texas Hold'Em No-Limit}
        \label{fig:hunl}
    \end{subfigure}
    \caption{Final performance of PPO-NFSP, PQN-NFSP, and ReBeL against pre-trained ReBeL evaluation opponent in Heads-Up Texas Hold'Em. Each algorithm is trained in self-play with 1e7 samples and evaluated across 10 seeds. Evaluation opponent is trained in self-play with 5e6 samples. Dashed lines indicate break-even performance.}
    \label{fig:holdem}
\end{figure}

In Figure~\ref{fig:holdem} we test the performance of baseline algorithms in Heads-Up (two-player) Texas Hold'Em Poker. In order to provide standardized performance results, we use ReBeL trained in self-play for 5e6 steps as an evaluation opponent. As expected, ReBeL performs well when tested against itself, due to access to more training samples and exploitation strategies learned during self-play. PPO-NFSP and PQN-NFSP are typically unable to reach break-even performance despite training on double the number of samples as the ReBeL baseline. This is likely due to ReBeL's well-tuned formulation for Hold'Em Poker, which utilizes both depth-limited CFR for subgame solving and function approximation, thereby leveraging game knowledge which is unavailable in model-free RL.

\begin{figure}[htbp]
    \centering
    \begin{subfigure}[b]{0.3\textwidth}
        \centering
        \includegraphics[width=\textwidth]{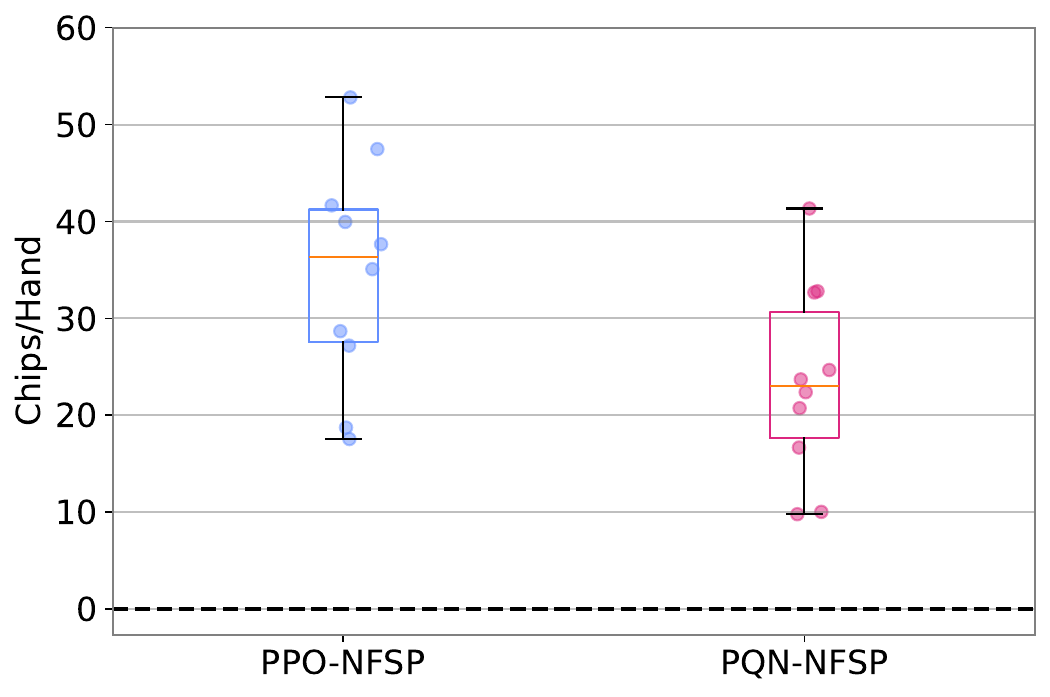}
        \caption{Five Card Draw}
        \label{fig:5cd}
    \end{subfigure}
    \hfill
    \begin{subfigure}[b]{0.3\textwidth}
        \centering
        \includegraphics[width=\textwidth]{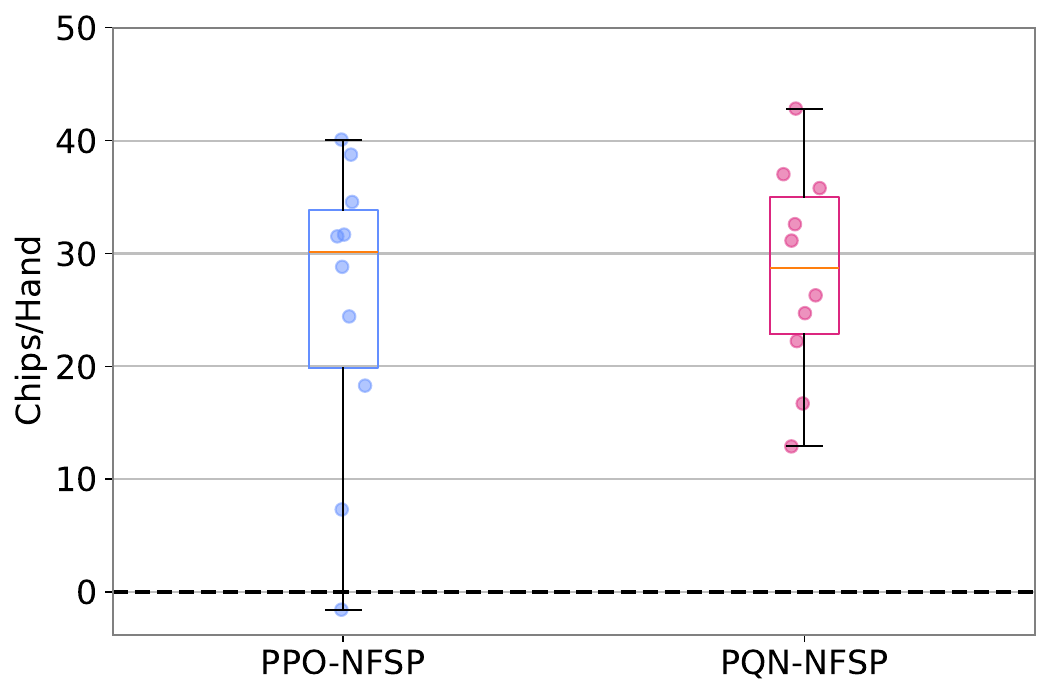}
        \caption{Seven Card Stud}
        \label{fig:7cs}
    \end{subfigure}
    \hfill
    \begin{subfigure}[b]{0.3\textwidth}
        \centering
        \includegraphics[width=\textwidth]{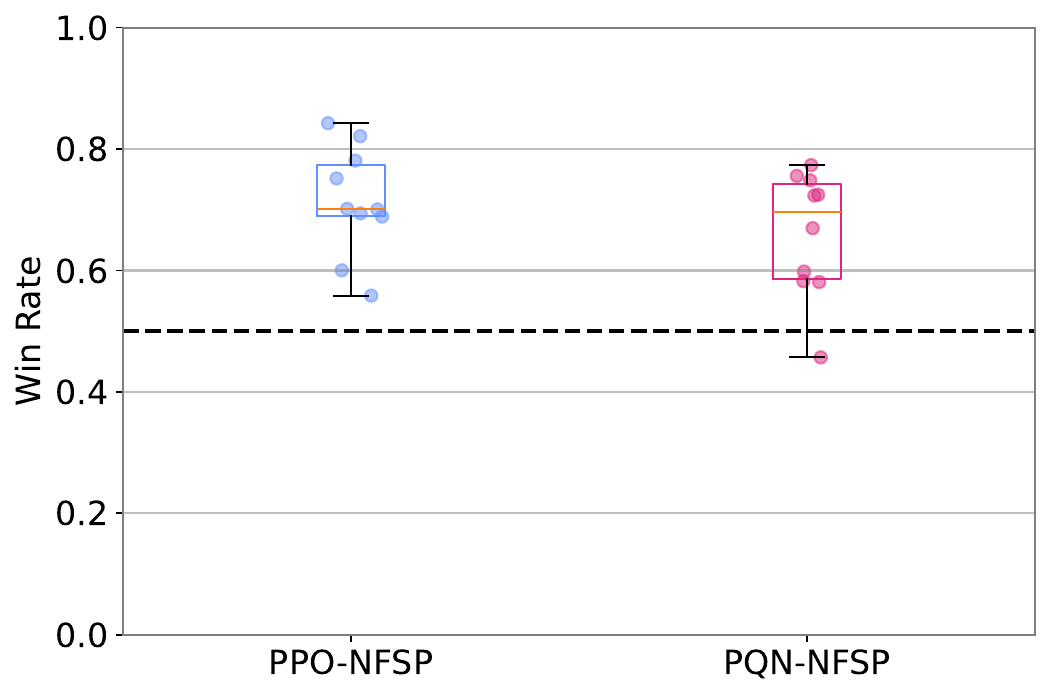}
        \caption{Werewolf}
        \label{fig:werewolf}
    \end{subfigure}

    \begin{subfigure}[b]{0.3\textwidth}
        \centering
        \includegraphics[width=\textwidth]{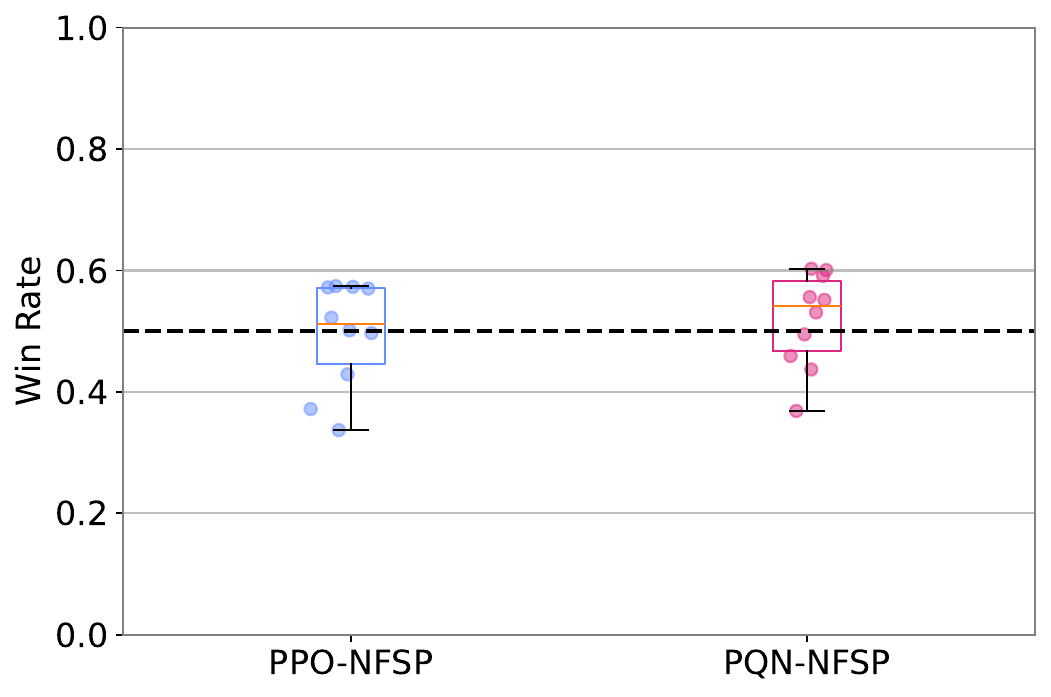}
        \caption{Goofspiel}
        \label{fig:goofspiel}
    \end{subfigure}
    \hfill
    \begin{subfigure}[b]{0.3\textwidth}
        \centering
        \includegraphics[width=\textwidth]{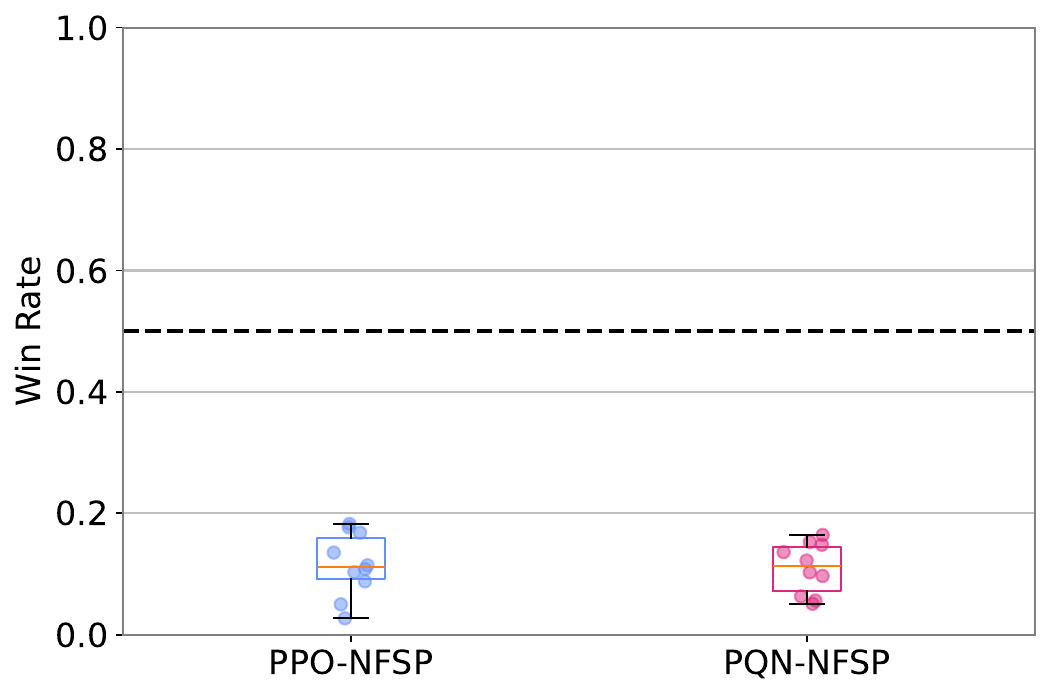}
        \caption{Kemps}
        \label{fig:kemps}
    \end{subfigure}
    \hfill
    \begin{subfigure}[b]{0.3\textwidth}
        \centering
        \includegraphics[width=\textwidth]{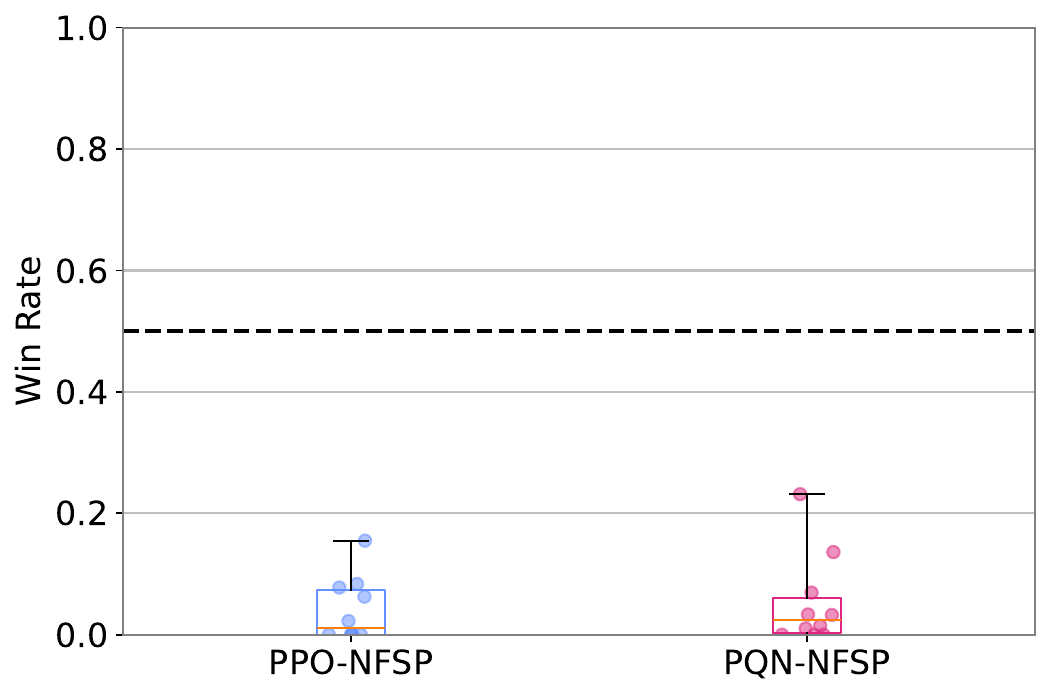}
        \caption{Bluff}
        \label{fig:bluff}
    \end{subfigure}
    \caption{Final performance of PPO-NFSP and PQN-NFSP against pre-trained PPO-NFSP evaluation opponent in BluffJAX games. Each algorithm is trained in self-play with 1e7 samples and evaluated across 10 seeds. Evaluation opponent is trained in self-play with 5e6 samples. Dashed lines indicate break-even performance.}
    \label{fig:other_games}
\end{figure}

Finally, in Figure~\ref{fig:other_games} we evaluate the performance of baseline algorithms in the remaining 6 games in BluffJAX. Note that we do not test ReBeL here as it utilizes model-based subgame solving, which is currently only available for Heads-Up Hold'Em variants. Therefore, we use PPO-NFSP trained in self-play for 5e6 steps as an evaluation opponent. Across the 6 games tested, PPO-NFSP had higher average performance in 3 with higher margins relative to PQN. Notably, the results for Five Card Draw, Seven Card Stud, and Werewolf indicate that the baselines were able to gain a noticeable advantage in certain games despite their lack of game knowledge exploitation. However, both algorithms struggled to gain an advantage in Goofspiel, likely because performing well requires a stochastic policy. Finally, both algorithms failed to break even in Kemps and Bluff. It should be noted that, by virtue of their game mechanics, these are the two environments in BluffJAX which do not enforce a game-ending condition in finite time (i.e. games can end via reaching a horizon). Thus, low win rates in these games are due to draws in horizon-timeouts. Kemps introduces difficult mechanics: not only must agents learn to signal when they have built a hand, but their partner must also correctly register that signal - certainly a strong challenge in protocol formation for RL. Similarly, Bluff requires introduces a harsh `theory of mind' challenge to adversarial RL, as agents must successfully predict when opponents are lying.

%% file: 01_sections/06_conclusion.tex
\section{Conclusion}
We introduce BluffJAX, an open-source suite of adversarial imperfect information games implemented in JAX for GPU-accelerated RL research, including five new games which have not been studied in RL thus far. We describe our API, design choices, and game implementations. We benchmark the performance of our implementations in terms of simulation throughput and memory usage in single and multi-GPU settings and compare against other libraries. We evaluate the performance of game-theoretic solvers, model-free RL, and model-based/tree search algorithms on our environments. We hope that BluffJAX will enable RL researchers to explore new challenges by providing fast and vectorized environments.

\textbf{Limitations and Future Work.} A key limitation is that BluffJAX is currently not equipped with PyTorch wrappers; therefore, it mostly caters to users already familiar with JAX and vectorized training. Additionally, the suite currently has limited performance evaluation tools. While solvers and baseline algorithms are provided for all games, a wider range of baselines would enable better standardization of results, such as external poker benchmarks. The sub-log-linear scaling of certain environments at higher parallelization (e.g. Kemps) suggests further optimizations can be made. Future plans include expanding the suite with environments such as Battleship, Mahjong, and Coup.

%% file: 01_sections/091_app_games.tex
\appendix
\section{Games in BluffJAX}\label{supp:games}
\subsection{Kuhn Poker}

\begin{minipage}{0.3\textwidth}
    \centering
    \includegraphics[width=\linewidth]{02_images/hero/kuhn.pdf}
    \label{fig:supp_kuhn}
\end{minipage}
\begin{minipage}{0.48\textwidth}
    \begin{center}
        \begin{tabular}{ll}
        \hline
            \multicolumn{1}{l}{Environment}  &\multicolumn{1}{l}{Kuhn Poker}\\ \hline
            Number of players (n) & 2\\
            Observation size & 9 \\
            Number of actions & 2 \\
            Number of JAX branches & 1\\
            Metric & Exploitability\\
            API Type & AEC\\
            \hline
        \end{tabular}
    \end{center}
\end{minipage}

\textbf{Objective:} Kuhn Poker is a well-studied two-player zero-sum game~\citep{kuhn1953extensive}. There are 3 cards: Jack, Queen, and King. At the start of a round, players are randomly assigned one private card each, and each player antes 1. Each player then gets one chance to bet. If player 1 raises, player 2 can either call or fold. If player 1 checks, player 2 can check or raise. If player 2 raises after player 1 checked, player 1 can either call or fold. If either player folds, the other player wins the pot. If neither player folds, there is a showdown and the higher card wins.

\textbf{Observations}
\begin{center}
    \begin{tabular}{lll}
    \hline
        \multicolumn{1}{l}{Input feature}  &\multicolumn{1}{l}{Feature Dimension} &\multicolumn{1}{l}{Feature type} \\ \hline
        Agent's private card & 3 & One-hot \\
        Self chips in pot & 2 & One-hot \\
        Opponent chips in pot & 2 & One-hot \\
        Player order (first or second to act) & 2 & One-hot \\
        \hline
        Total & 9 \\
        \hline
    \end{tabular}
\end{center}

\textbf{Actions}
\begin{center}
    \begin{tabular}{lll}
    \hline
        \multicolumn{1}{l}{Action}  &\multicolumn{1}{l}{Logit} \\ \hline
        Check/fold & 0  \\
        Raise/call & 1  \\
        \hline
        Total & 2 \\
        \hline
    \end{tabular}
\end{center}

\textbf{Rewards:} Agents receive net chips gained as terminal rewards.

\subsection{Leduc Hold'em}

\begin{minipage}{0.3\textwidth}
    \centering
    \includegraphics[width=\linewidth]{02_images/hero/leduc.pdf}
    \label{fig:supp_leduc}
\end{minipage}
\begin{minipage}{0.48\textwidth}
    \begin{center}
        \begin{tabular}{ll}
        \hline
            Environment & Leduc Hold'em\\ \hline
            Number of players (n) & 2\\
            Observation size & 36 \\
            Number of actions & 3 \\
            Number of JAX branches & 3\\
            Metric & Exploitability\\
            API Type & AEC\\
            \hline
        \end{tabular}
    \end{center}
\end{minipage}

\textbf{Objective:} Two-player poker with a 6-card deck (two J, two Q, two K), one private card per player, and one public card revealed after round 1. Players bet over two rounds. Best hand (pair beats high card) wins at showdown, or a fold ends the hand early.

\textbf{Observations}
\begin{center}
    \begin{tabular}{lll}
    \hline
        Input feature & Feature Dimension & Feature type \\ \hline
        Agent private rank (J/Q/K) & 3 & One-hot \\
        Public rank (if revealed) & 3 & One-hot \\
        Self ante bucket (0--14+) & 15 & One-hot \\
        Opponent ante bucket (0--14+) & 15 & One-hot \\
        \hline
        Total & 36 & \\
        \hline
    \end{tabular}
\end{center}

\textbf{Actions}
\begin{center}
    \begin{tabular}{ll}
    \hline
        Action & Logit \\ \hline
        Fold & 0 \\
        Call/Check & 1 \\
        Raise & 2 \\
        \hline
        Total & 3 \\
        \hline
    \end{tabular}
\end{center}

\textbf{Rewards:}
Terminal reward is net chips relative to starting stack (non-terminal reward is 0).

\subsection{Texas Limit Hold'em}

\begin{minipage}{0.3\textwidth}
    \centering
    \includegraphics[width=\linewidth]{02_images/hero/tul.pdf}
    \label{fig:supp_tul}
\end{minipage}
\begin{minipage}{0.48\textwidth}
    \begin{center}
        \begin{tabular}{ll}
        \hline
            Environment & Texas Limit Hold'em\\ \hline
            Number of players (n) & 2--10 (common setting)\\
            Observation size & 72 + n \\
            Number of actions & 4 \\
            Number of JAX branches & 4\\
            Metric & Chips/Hand\\
            API Type & AEC\\
            \hline
        \end{tabular}
    \end{center}
\end{minipage}

\textbf{Objective:} Multi-player limit hold'em with fixed betting amounts and blinds. Players maximize expected chip gain by betting/folding through preflop, flop, turn, and river.

\textbf{Observations}
\begin{center}
    \begin{tabular}{lll}
    \hline
        Input feature & Feature Dimension & Feature type \\ \hline
        Private + visible public cards & 52 & Binary \\
        Raise-count history (4 rounds, 0-4 raises each) & 20 & One-hot \\
        Position relative to small blind & n & One-hot \\
        \hline
        Total & 72 + n & \\
        \hline
    \end{tabular}
\end{center}

\textbf{Actions}
\begin{center}
    \begin{tabular}{ll}
    \hline
        Action & Logit \\ \hline
        Call & 0 \\
        Raise & 1 \\
        Fold & 2 \\
        Check & 3 \\
        \hline
        Total & 4 \\
        \hline
    \end{tabular}
\end{center}

\textbf{Rewards:}
Rewards are terminal chip payoffs normalized by big blind.

\subsection{Texas No-Limit Hold'em}

\begin{minipage}{0.3\textwidth}
    \centering
    \includegraphics[width=\linewidth]{02_images/hero/tunl.pdf}
    \label{fig:supp_tunl}
\end{minipage}
\begin{minipage}{0.48\textwidth}
    \begin{center}
        \begin{tabular}{ll}
        \hline
            Environment & Texas No-Limit Hold'em\\ \hline
            Number of players (n) & 2-10 (common setting)\\
            Observation size & 54 \\
            Number of actions & 5 \\
            Number of JAX branches & 5\\
            Metric & Chips/Hand\\
            API Type & AEC\\
            \hline
        \end{tabular}
    \end{center}
\end{minipage}

\textbf{Objective:} No-limit Hold'em variant with discrete raise abstractions (half-pot, pot, all-in). Agents optimize chip profit over each hand.

\textbf{Observations}
\begin{center}
    \begin{tabular}{lll}
    \hline
        Input feature & Feature Dimension & Feature type \\ \hline
        Private + visible public cards & 52 & Binary \\
        Current player's chips in pot & 1 & Scalar \\
        Max chips in pot among players & 1 & Scalar \\
        \hline
        Total & 54 & \\
        \hline
    \end{tabular}
\end{center}

\textbf{Actions}
\begin{center}
    \begin{tabular}{ll}
    \hline
        Action & Logit \\ \hline
        Check/Call & 0 \\
        Raise half-pot & 1 \\
        Raise pot & 2 \\
        All-in & 3 \\
        Fold & 4 \\
        \hline
        Total & 5 \\
        \hline
    \end{tabular}
\end{center}

\textbf{Rewards:}
Rewards are terminal net chips gained (chips won minus chips invested).

\subsection{5 Card Draw}

\begin{minipage}{0.3\textwidth}
    \centering
    \includegraphics[width=\linewidth]{02_images/hero/five_card_draw.pdf}
    \label{fig:supp_five}
\end{minipage}
\begin{minipage}{0.48\textwidth}
    \begin{center}
        \begin{tabular}{ll}
        \hline
            Environment & 5 Card Draw\\ \hline
            Number of players (n) & 2--10\\
            Observation size & 54 \\
            Number of actions & 37 \\
            Number of JAX branches & 11\\
            Metric & Chips/Hand\\
            API Type & AEC\\
            \hline
        \end{tabular}
    \end{center}
\end{minipage}

\textbf{Objective:} Players play one draw-poker hand with a betting round, a draw phase, and a final betting round, maximizing expected chip gain.

\textbf{Observations}
\begin{center}
    \begin{tabular}{lll}
    \hline
        Input feature & Feature Dimension & Feature type \\ \hline
        Current player 5-card hand (in 52-card index space) & 52 & Binary \\
        Current player's chips in pot & 1 & Scalar \\
        Max chips in pot among players & 1 & Scalar \\
        \hline
        Total & 54 & \\
        \hline
    \end{tabular}
\end{center}

\textbf{Actions}
\begin{center}
    \begin{tabular}{ll}
    \hline
        Action & Logit \\ \hline
        Check/Call, Raise 1/2 Pot, Raise Pot, All-in, Fold & 0--4 \\
        Draw pattern (keep/discard over 5 cards) & 5--36 \\
        \hline
        Total & 37 \\
        \hline
    \end{tabular}
\end{center}

\textbf{Rewards:}
Rewards are terminal net chips gained.

\subsection{7 Card Stud}

\begin{minipage}{0.3\textwidth}
    \centering
    \includegraphics[width=\linewidth]{02_images/hero/seven_card_stud.pdf}
    \label{fig:supp_seven}
\end{minipage}
\begin{minipage}{0.48\textwidth}
    \begin{center}
        \begin{tabular}{ll}
        \hline
            Environment & 7 Card Stud\\ \hline
            Number of players (n) & 2--10\\
            Observation size & $52 + 208(n-1) + 25 + n$ \\
            Number of actions & 4 \\
            Number of JAX branches & 4\\
            Metric & Chips/Hand\\
            API Type & AEC\\
            \hline
        \end{tabular}
    \end{center}
\end{minipage}

\textbf{Objective:} Stud poker with antes/bring-in and five betting streets. Players maximize expected chip return by betting and showing down strongest 7-card hand among active players.

\textbf{Observations}
\begin{center}
    \begin{tabular}{lll}
    \hline
        Input feature & Feature Dimension & Feature type \\ \hline
        Own 7 cards (global deck index) & 52 & Binary \\
        Opponents' visible upcards (4 slots each opponent) & $208(n-1)$ & One-hot blocks \\
        Raise-count history (5 rounds, 0--4) & 25 & One-hot \\
        Position relative to bring-in & n & One-hot \\
        \hline
        Total & $52 + 208(n-1) + 25 + n$ & \\
        \hline
    \end{tabular}
\end{center}

\textbf{Actions}
\begin{center}
    \begin{tabular}{ll}
    \hline
        Action & Logit \\ \hline
        Call & 0 \\
        Raise & 1 \\
        Fold & 2 \\
        Check & 3 \\
        \hline
        Total & 4 \\
        \hline
    \end{tabular}
\end{center}

\textbf{Rewards:}
Terminal payoff is net chips normalized by big bet.

\subsection{Goofspiel}

\begin{minipage}{0.3\textwidth}
    \centering
    \includegraphics[width=\linewidth]{02_images/hero/goofspiel.pdf}
    \label{fig:supp_goof}
\end{minipage}
\begin{minipage}{0.48\textwidth}
    \begin{center}
        \begin{tabular}{ll}
        \hline
            Environment & Goofspiel\\ \hline
            Number of players (n) & 2\\
            Observation size & 39 \\
            Number of actions & 13 \\
            Number of JAX branches & 1\\
            Metric & Return/Hand (or Win Rate)\\
            API Type & Parallel\\
            \hline
        \end{tabular}
    \end{center}
\end{minipage}

\textbf{Objective:} Both players bid cards simultaneously for sequential prize cards. Highest unique bid wins prize points; ties discard the prize.

\textbf{Observations}
\begin{center}
    \begin{tabular}{lll}
    \hline
        Input feature & Feature Dimension & Feature type \\ \hline
        Current prize card & 13 & One-hot \\
        Self cards already used & 13 & Binary \\
        Opponent cards already used & 13 & Binary \\
        \hline
        Total & 39 & \\
        \hline
    \end{tabular}
\end{center}

\textbf{Actions}
\begin{center}
    \begin{tabular}{ll}
    \hline
        Action & Logit \\ \hline
        Play bid card (rank 1--13) & 0--12 \\
        \hline
        Total & 13 \\
        \hline
    \end{tabular}
\end{center}

\textbf{Rewards:}
Per-round reward is prize value to the unique highest bidder; terminal return is cumulative points.

\subsection{Bluff}

\begin{minipage}{0.3\textwidth}
    \centering
    \includegraphics[width=\linewidth]{02_images/hero/bluff.pdf}
    \label{fig:supp_bluff}
\end{minipage}
\begin{minipage}{0.48\textwidth}
    \begin{center}
        \begin{tabular}{ll}
        \hline
            Environment & Bluff\\ \hline
            Number of players (n) & $\geq 3$\\
            Observation size & $5 \cdot \text{deck\_size} + 3$ (default 263) \\
            Number of actions & 13 \\
            Number of JAX branches & 10\\
            Metric & Win Rate\\
            API Type & AEC\\
            \hline
        \end{tabular}
    \end{center}
\end{minipage}

\textbf{Objective:} Turn-based social bluffing game (Cheat/I Doubt It). Players make claims, play hidden cards, and other players may challenge. Goal is to empty your hand first.

\textbf{Observations}
\begin{center}
    \begin{tabular}{lll}
    \hline
        Input feature & Feature Dimension & Feature type \\ \hline
        Pile claims & $52$ & Thermometer \\
        Pile size & $52$ & Thermometer \\
        Own hand counts & $52$ & Thermometer \\
        Claimant hand size (challenge phase) & $52$ & Thermometer \\
        Current claim size (challenge phase) & $52$ & Thermometer \\
        Phase id (claim/play/challenge) & 3 & One-hot \\
        \hline
        Total & $5 \cdot \text{deck\_size} + 3$ & \\
        \hline
    \end{tabular}
\end{center}

\textbf{Actions}
\begin{center}
    \begin{tabular}{ll}
    \hline
        Action & Logit \\ \hline
        Phase-dependent action index & 0--12 \\
        \hline
        Total & 13 \\
        \hline
    \end{tabular}
\end{center}

\textbf{Rewards:} Shaped rewards are used for successful/failed challenges and card outcomes; winner also receives terminal bonus.

\subsection{Werewolf}

\begin{minipage}{0.3\textwidth}
    \centering
    \includegraphics[width=\linewidth]{02_images/hero/werewolf.pdf}
    \label{fig:supp_werewolf}
\end{minipage}
\begin{minipage}{0.48\textwidth}
    \begin{center}
        \begin{tabular}{ll}
        \hline
            Environment & Werewolf\\ \hline
            Number of players (n) & 6 (current implementation)\\
            Observation size & $7 + 6n$ \\
            Number of actions & $n + 1$ \\
            Number of JAX branches & 7\\
            Metric & Win Rate\\
            API Type & AEC\\
            \hline
        \end{tabular}
    \end{center}
\end{minipage}

\textbf{Objective:} Hidden-role social deduction game with phases Night $\rightarrow$ Accuse $\rightarrow$ Vote. Humans try to eliminate all werewolves; werewolves try to reach parity with humans.

\textbf{Observations}
\begin{center}
    \begin{tabular}{lll}
    \hline
        Input feature & Feature Dimension & Feature type \\ \hline
        Own role & 4 & One-hot \\
        Current phase & 3 & One-hot \\
        Alive players (relative indexing) & n & Binary \\
        Werewolf teammate info (if werewolf) & n & Binary \\
        Seer belief/results (if seer) & $3n$ & Ternary one-hot per player \\
        Accusation targets this round & n & Binary \\
        \hline
        Total & $7 + 6n$ & \\
        \hline
    \end{tabular}
\end{center}

\textbf{Actions}
\begin{center}
    \begin{tabular}{ll}
    \hline
        Action & Logit \\ \hline
        Relative target player index & $0 \ldots n-1$ \\
        No-op & $n$ \\
        \hline
        Total & $n+1$ \\
        \hline
    \end{tabular}
\end{center}

\textbf{Rewards:} At terminal state, all winners get +10 and all losers get -10.

\subsection{Kemps}

\begin{minipage}{0.3\textwidth}
    \centering
    \includegraphics[width=\linewidth]{02_images/hero/kemps.pdf}
    \label{fig:supp_kemps}
\end{minipage}
\begin{minipage}{0.48\textwidth}
    \begin{center}
        \begin{tabular}{ll}
        \hline
            Environment & Kemps\\ \hline
            Number of players (n) & 4,6,8,\dots (even)\\
            Observation size & $52(n+1) + n \cdot \text{comm}$ (default deck) \\
            Number of actions & $172 \cdot \text{comm}$ \\
            Number of JAX branches & 1\\
            Metric & Win Rate\\
            API Type & Parallel\\
            \hline
        \end{tabular}
    \end{center}
\end{minipage}

\textbf{Objective:} Partnership card game with simultaneous moves. Teams try to form four-of-a-kind and correctly declare KEMPS (or STOP KEMPS against opponents).

\textbf{Observations}
\begin{center}
    \begin{tabular}{lll}
    \hline
        Input feature & Feature Dimension & Feature type \\ \hline
        Private hand & $52$ & Binary \\
        Public center cards & 52 & Binary \\
        Relative communication signals (all players) & $n \cdot \text{comm}$ & One-hot \\
        \hline
        Total & $52(n+1) + n \cdot \text{comm}$ & \\
        \hline
    \end{tabular}
\end{center}

\textbf{Actions}
\begin{center}
    \begin{tabular}{ll}
    \hline
        Action & Logit \\ \hline
        Rank swap (lose-rank, gain-rank) & $0 \ldots 168$ \\
        No-op & 169 \\
        Declare KEMPS & 170 \\
        Declare STOP KEMPS & 171 \\
        Each paired with communication signal & $\times \text{comm}$ \\
        \hline
        Total & $172 \cdot \text{comm}$ \\
        \hline
    \end{tabular}
\end{center}

\textbf{Rewards:} Terminal team reward is +1/-1 based on declaration correctness or correctness of spotting opposing team's signal

%% file: 01_sections/092_app_design.tex
\section{Design Choices}\label{supp:design}
In this section, we hope to motivate some of the design choices we made in order to optimize BluffJAX for performance and usability.

\subsection{Binary or One-Hot Features?}
Our environments often contain features which are binary, e.g. whether the agent currently has the ability to play a certain action or not. We conduct a simple test to see whether a neural network is better able to distinguish features of this type when they are represented in binary format [0/1] or one-hot format [10/01]. We train a neural network on a randomized regression task with one input feature and two output features. We train the network on 20,000 examples over 10 epochs. In order to standardize the number of input neurons, we feed the binary features as [00/11] instead of [0/1].

\begin{figure}[htbp]
    \centering
    \begin{subfigure}[b]{0.45\textwidth}
        \centering
        \includegraphics[width=\textwidth]{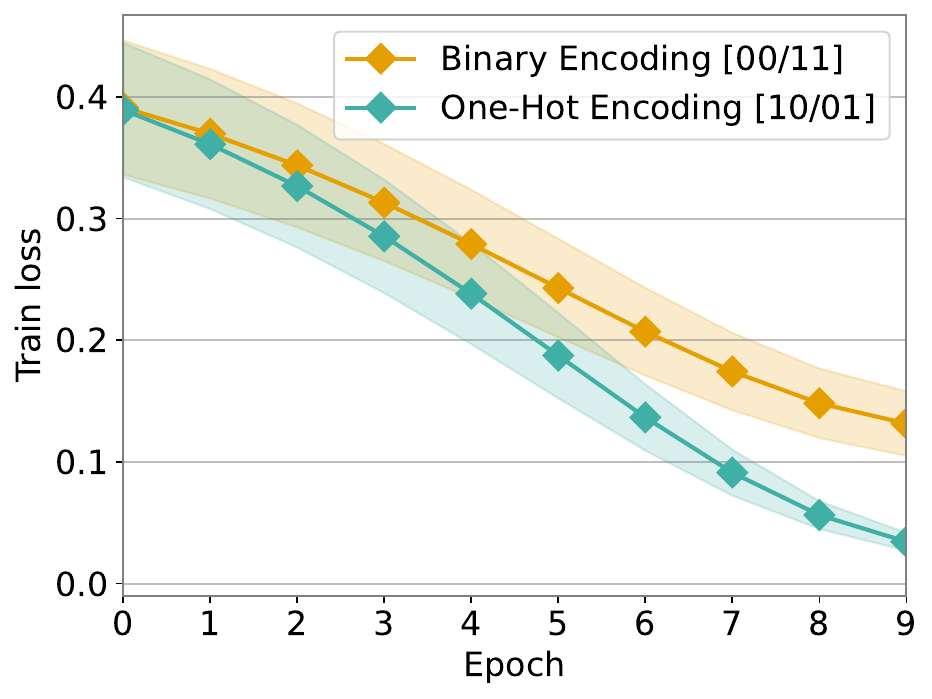}
        \caption{Training loss}
        \label{fig:loss}
    \end{subfigure}
    \hfill
    \begin{subfigure}[b]{0.45\textwidth}
        \centering
        \includegraphics[width=\textwidth]{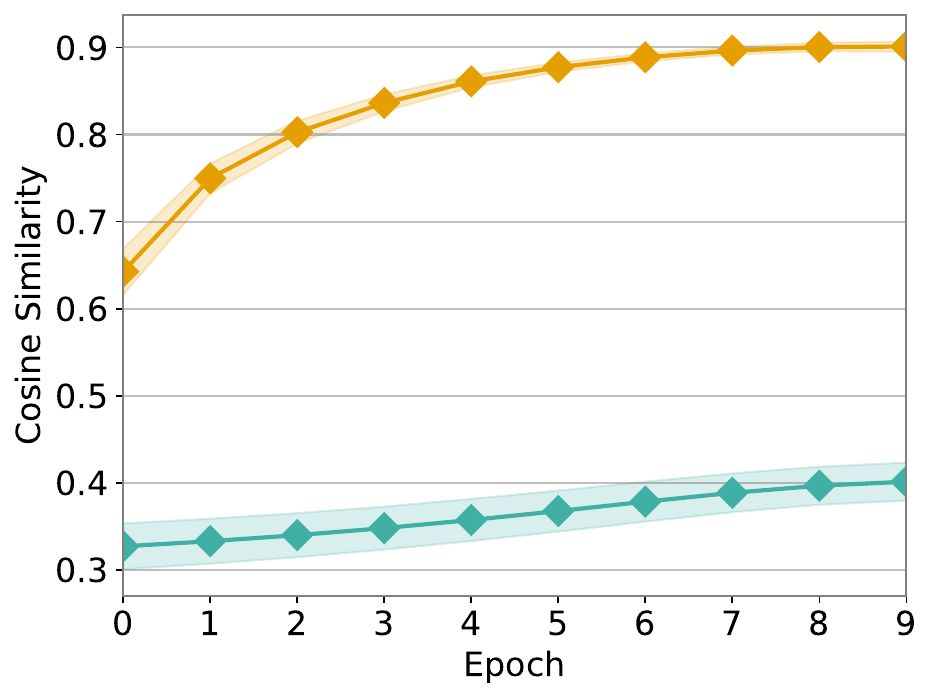}
        \caption{Representation cosine similarity}
        \label{fig:css}
    \end{subfigure}
    \caption{Training loss and cosine similarity of representations during training. Results show mean and std error over 10 seeds with 20,000 datapoints each.}
    \label{fig:logits}
\end{figure}

Figure~\ref{fig:logits} shows the results; the training loss indicates that the network learns the regression task faster when the features are one-hot rather than binary. This makes intuitive sense, as one-hot features `separate' the distinct input types, thereby saving the network from having to build an internal representation to partition `0' inputs from `1' inputs. However, Figure~\ref{fig:css} demonstrates that the cosine similarity of the internal representations of the distinct inputs was \textit{already} low at the start of training (inferring that they were already separated in feature space) and they actually become more similar over the course of training. This suggests that one-hot features can help learning when used, but their representational separation is not necessarily optimized for or maintained during learning.

\subsection{Parallel or AEC API?}
One consideration when designing a multi-agent benchmark is whether the environments are more amenable to a parallel agent-environment interaction interface or an agent-environment-cycle (AEC) interface. Similar to PettingZoo~\citep{terry2021pettingzoo}, we choose to implement both in BluffJAX. In turn-based games where most steps only involve one agent, a parallel interface wastes significant compute by querying all other agents but forcing them to choose noop actions; on the other hand, certain environments like Goofspiel naturally follow a parallel-action game loop.

\subsection{Handle resets internally or externally?}
In parallelized settings, JAX requires all conditional logic to be executed regardless of whether the results are used or not\footnote{https://docs.jax.dev/en/latest/\_autosummary/jax.lax.cond.html}. Therefore, the typical RL `state$\rightarrow$action$\rightarrow$ next state' loop must be accompanied at each instance by reset logic, and the correct `next state' is chosen depending on the termination condition of the episode. The choice thus remains of whether to handle this logic inside the environment API, or to expose this logic for the user to handle. In BluffJAX, we choose to handle this logic internally, as users don't typically need to know/worry about this, and exposing it simply adds to the confusion that befalls JAX newcomers.

\subsection{Game logic datatypes: floats or ints?}
This question arises as a consequence of conversions; internal game logic must often convert actions (which arrive as \mintinline{Python}{jax.numpy.int32} in discrete settings), and the input to a neural network must typically be of the type \mintinline{Python}{jax.numpy.float32}. Therefore, which datatype should be used for internal game logic? Does the conversion cost matter? We find that the cost is generally \textbf{negligible} within the overall training loop, and therefore we simply use whichever datatype is most convenient in each case.

\subsection{Done, absorbing, truncated, terminated?}
Environment suites do not generally agree on a convention on how to distinguish when agents reach absorbing states or reach a horizon. We choose to implement both: our API provides a boolean array \mintinline{Python}{absorbing} to indicate the absorbing state of each agent (as some may drop out of a game earlier than others) as well as a single boolean \mintinline{Python}{done} if all agents have reached absorbing states or if the global episode horizon has been reached.

\subsection{Type hints?}
What type hints should be given to users to understand the code? We find no universally consistent typing scheme for JAX functions and arrays. We therefore implement our own type hints that we hope are useful in distinguishing normal Python types from JAX arrays and pytrees.

\subsection{Randomize the start player?}
In self-play settings, this question has no bearing since the same network will be trained on all agent data. However, if distinct players are loaded into a game (e.g. if a training model is loaded as `player 1' and a baseline model as `player 2', then player order absolutely makes a difference, and thus we randomize the starting player in all cases for simplicity.

\subsection{Give the agent an observation of its own order?}
We find that this feature either does not help or actively harms learning in practice, as it causes agents to overfit to this feature rather than generalize across experiences while agnostic to the player they originated from. We therefore omit this feature in all games except where it explicitly matters (e.g. whether you are the start player or second player in Kuhn and Leduc Poker matters).

%% file: 01_sections/093_app_baselines.tex
\section{Details on Experiments}\label{supp:baselines}
In this section, we detail the hyperparameters and training settings for the experiments in the main paper.

\subsection{CFR and Deep CFR}
Our results for CFR were run with 100 iterations over the whole game tree for both games. Our results for Deep CFR were run with 50 iterations and 100 traversals per iteration. The policy network and advantage networks used in both cases are fully connected MLPs with 2 hidden layers, 64 hidden units/layer, ReLu activations, and LayerNorm after each hidden layer. The networks were trained with 200 advantage training steps and a memory capacity of 100,000. All networks were trained with the Adam~\citep{kingma2014adam} optimizer. We ran 10 seeds per game for Deep CFR.

\subsection{PPO, PQN, PPO-NFSP, PQN NFSP}
We benchmarked PPO and PPO-NFSP in all environments using the following settings for the value network and policy networks of PPO: fully connected MLPs with 2 hidden layers, 128 hidden units/layer, ReLu activations. Table~\ref{tab:ppo_hyperparams} details the hyperparameters and training settings for the PPO actor and value networks.

\begin{table}[htbp]
    \caption{PPO hyperparameters.}
    \begin{center}
        \begin{tabular}{lc}
        \hline
            \multicolumn{1}{c}{Hyperparameter} & \multicolumn{1}{c}{Value} \\ \hline
            num\_timesteps & $1e7$ \\
            num\_envs & 1024 \\
            num\_steps\_per\_env\_per\_update & 64 \\
            num\_epochs & 4 \\
            num\_minibatches & 4 \\
            lr & $3\times10^{-5}$ \\
            anneal\_lr & True \\
            gamma & 0.99 \\
            gae\_lambda & 0.95 \\
            clip\_eps & 0.2 \\
            vf\_clip & 0.2 \\
            ent\_coef & 0.0 \\
            vf\_coef & 1.0 \\
            max\_grad\_norm & 0.5 \\
            optimizer & adam \\
            \hline
        \end{tabular}
    \end{center}
    \label{tab:ppo_hyperparams}
\end{table}

We also benchmarked PQN and PQN-NFSP in all environments. Here are the settings for the PQN networks in each case: fully connected MLPs with 2 hidden layers, 128 hidden units/layer, LayerNorm after each layer, ReLu activations. Table~\ref{tab:pqn_hyperparams} details the hyperparameters and training settings for the PQN Q-network.

\begin{table}[htbp]
    \caption{PQN hyperparameters.}
    \begin{center}
        \begin{tabular}{lc}
        \hline
            \multicolumn{1}{c}{Hyperparameter} & \multicolumn{1}{c}{Value} \\ \hline
            num\_timesteps & $1e7$ \\
            num\_envs & 1024 \\
            num\_steps\_per\_env\_per\_update & 64 \\
            num\_epochs & 2 \\
            num\_minibatches & 2 \\
            lr & $2.5\times10^{-4}$ \\
            start\_e & 1.0 \\
            end\_e & 0.05 \\
            exploration\_fraction & 0.2 \\
            q\_lambda & 0.9 \\
            max\_grad\_norm & 0.5 \\
            gamma & 0.99 \\
            fc\_dim\_size & 128 \\
            anticipatory\_eta & 0.1 \\
            sl\_reservoir\_capacity & 100000 \\
            sl\_batch\_size & 256 \\
            sl\_num\_steps\_per\_update & 2 \\
            sl\_lr & $3\times10^{-4}$ \\
            optimizer & adam \\
            \hline
        \end{tabular}
    \end{center}
    \label{tab:pqn_hyperparams}
\end{table}

In experiments with PPO-NFSP and PQN-NFSP, we use the settings above for the best response PPO/PQN agents. For the NFSP anticipatory (supervised) networks, we use the following settings: 2 hidden layers, 128 hidden units/layer, ReLu activations. Table~\ref{tab:nfsp_hyperparams} details the hyperparameters for the supervised network.

\begin{table}[htbp]
    \caption{NFSP Anticipatory Network hyperparameters.}
    \begin{center}
        \begin{tabular}{lc}
        \hline
            \multicolumn{1}{c}{Hyperparameter} & \multicolumn{1}{c}{Value} \\ \hline
            anticipatory\_eta & 0.1 \\
            sl\_reservoir\_capacity & 100000 \\
            sl\_batch\_size & 256 \\
            sl\_num\_steps\_per\_update & 2 \\
            sl\_lr & $3\times10^{-4}$ \\
            optimizer & adam \\
            \hline
        \end{tabular}
    \end{center}
    \label{tab:nfsp_hyperparams}
\end{table}

\subsection{ReBeL}
Our experiments with ReBeL had the following settings for the value network in each case: fully connected MLP with 2 hidden layers, 128 hidden units/layer, ReLu activations. Table~\ref{tab:rebel_hyperparams} details the hyperparameters and training settings for ReBeL.

\begin{table}[htbp]
    \caption{ReBeL hyperparameters.}
    \begin{center}
        \begin{tabular}{lc}
        \hline
            \multicolumn{1}{c}{Hyperparameter} & \multicolumn{1}{c}{Value} \\ \hline
            num\_update\_steps & 1000 \\
            num\_subgames\_per\_update & 2 \\
            cfr\_iterations & 2 \\
            max\_depth & 4 \\
            random\_action\_prob & 0.2 \\
            value\_lr & $3\times10^{-4}$ \\
            replay\_capacity & 10000 \\
            batch\_size & 256 \\
            num\_value\_train\_steps & 4 \\
            \hline
        \end{tabular}
    \end{center}
    \label{tab:rebel_hyperparams}
\end{table}

%% file: 01_sections/094_app_head2head.tex
\section{Head-to-Head Evaluations}\label{supp:head2head}

\begin{figure}[htbp]
    \centering
    \begin{subfigure}[b]{0.32\textwidth}
        \centering
        \includegraphics[width=\textwidth]{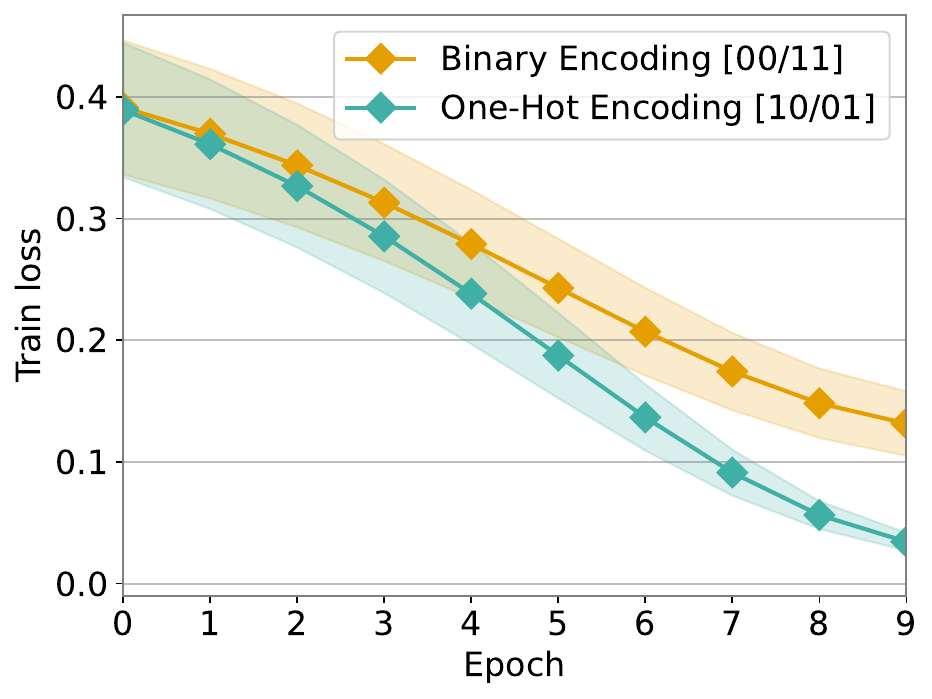}
        \caption{Training loss}
        \label{fig:loss}
    \end{subfigure}
    \hfill
    \begin{subfigure}[b]{0.32\textwidth}
        \centering
        \includegraphics[width=\textwidth]{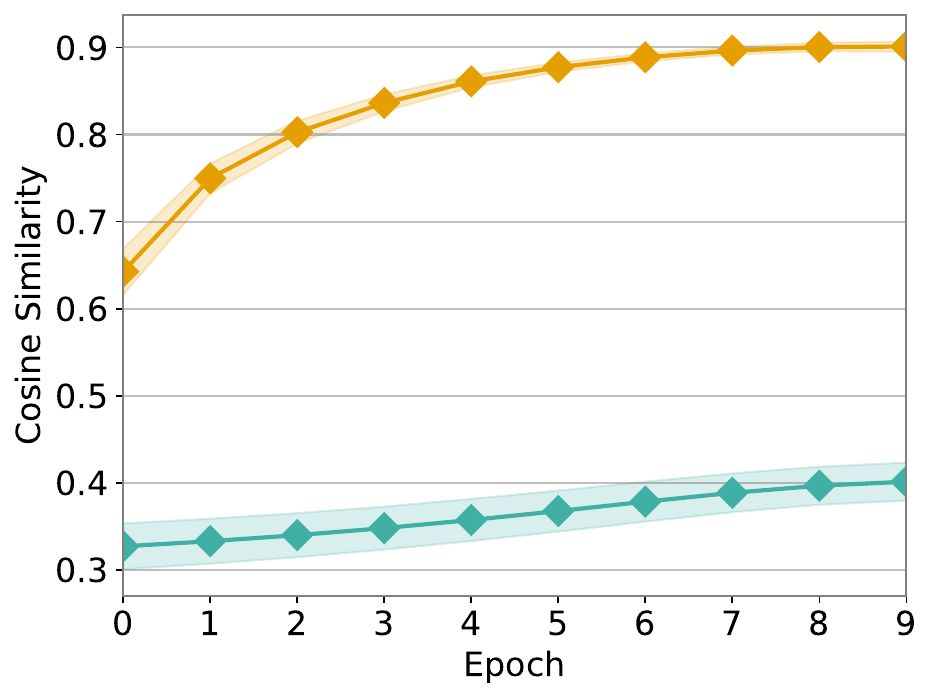}
        \caption{Representation cosine similarity}
        \label{fig:css}
    \end{subfigure}
    \hfill
    \begin{subfigure}[b]{0.32\textwidth}
        \centering
        \includegraphics[width=\textwidth]{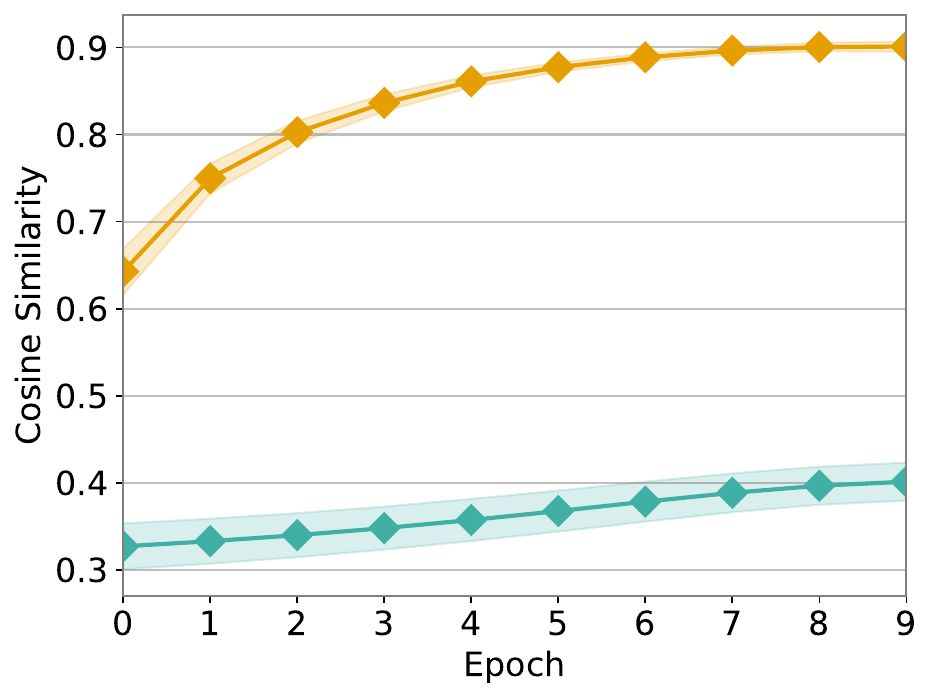}
        \caption{Representation cosine similarity}
        \label{fig:css}
    \end{subfigure}

    \vspace{0.5em}

    \begin{subfigure}[b]{0.32\textwidth}
        \centering
        \includegraphics[width=\textwidth]{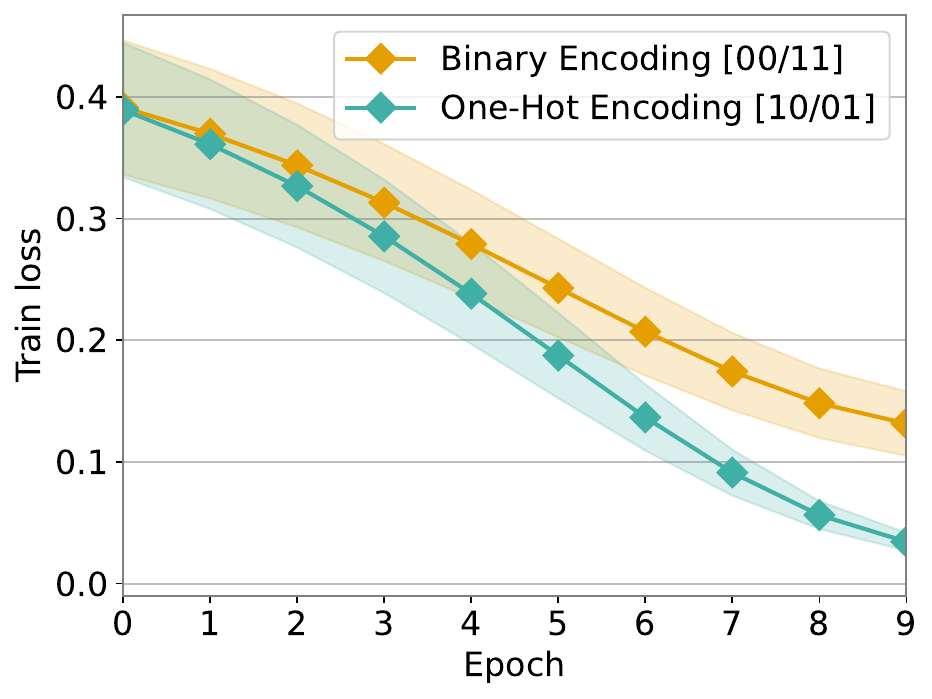}
        \caption{Training loss}
        \label{fig:loss}
    \end{subfigure}
    \hfill
    \begin{subfigure}[b]{0.32\textwidth}
        \centering
        \includegraphics[width=\textwidth]{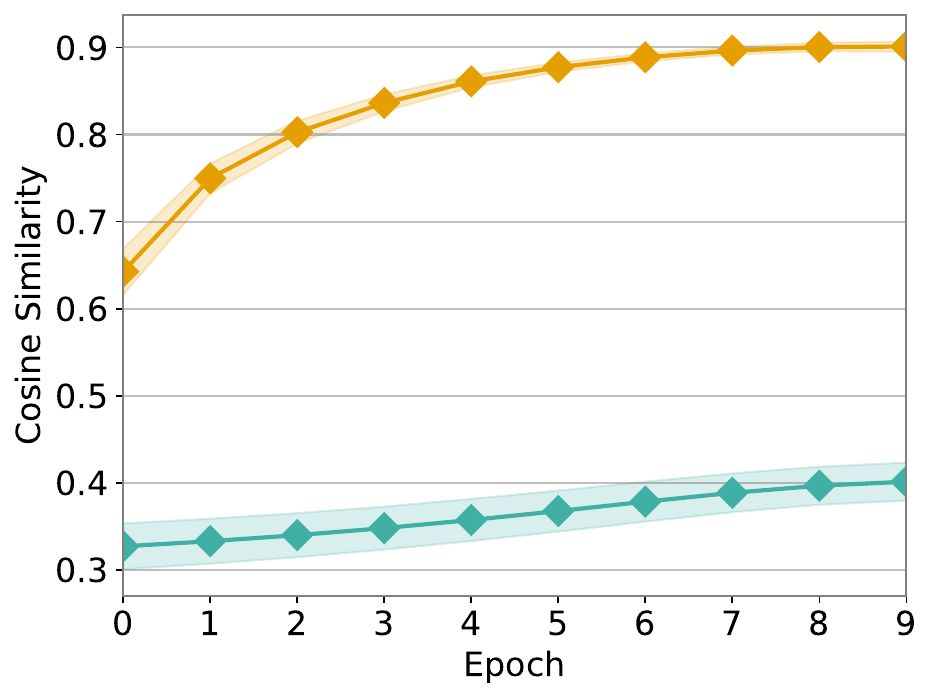}
        \caption{Representation cosine similarity}
        \label{fig:css}
    \end{subfigure}
    \hfill
    \begin{subfigure}[b]{0.32\textwidth}
        \centering
        \includegraphics[width=\textwidth]{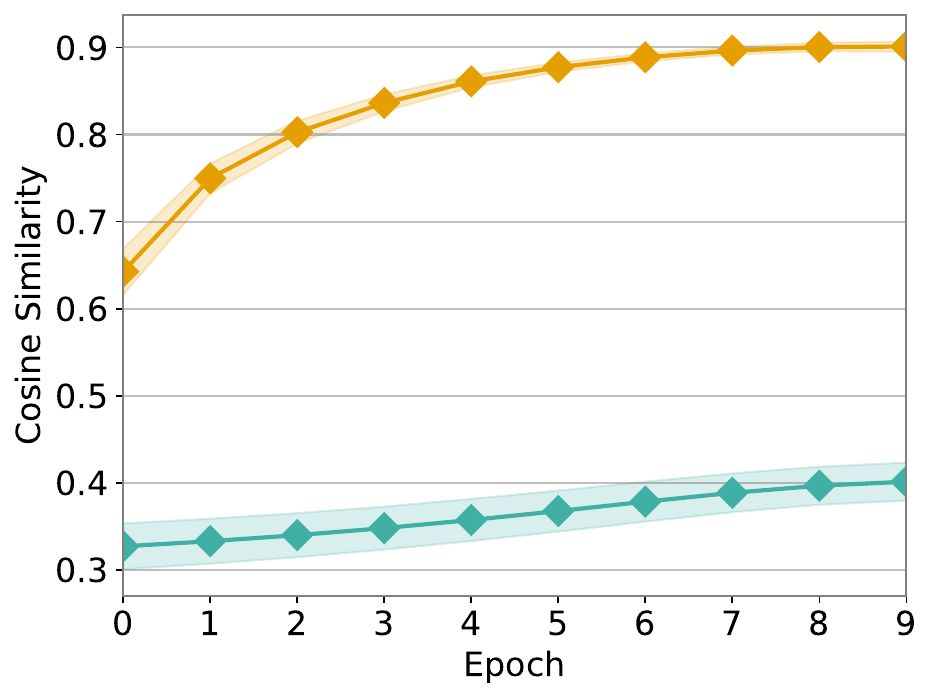}
        \caption{Representation cosine similarity}
        \label{fig:css}
    \end{subfigure}

    \caption{Training dynamics across twelve settings. Top row: training loss curves; bottom row: representation cosine similarity over training.}
    \label{fig:training_dynamics_grid}
\end{figure}

%% file: checklist.tex
\section*{NeurIPS Paper Checklist}

\begin{enumerate}

\item {\bf Claims}
    \item[] Question: Do the main claims made in the abstract and introduction accurately reflect the paper's contributions and scope?
    \item[] Answer: \answerYes{} 
    \item[] Justification: Our claims include: the open-source implementation of adversarial imperfect information games (supported by the code provided), high performance throughput on GPU devices (evidenced by Section~\ref{sec:benchmarking}), and baseline algorithmic results for future comparisons (evidenced by Section~\ref{sec:experiments}).
    \item[] Guidelines:
    \begin{itemize}
        \item The answer \answerNA{} means that the abstract and introduction do not include the claims made in the paper.
        \item The abstract and/or introduction should clearly state the claims made, including the contributions made in the paper and important assumptions and limitations. A \answerNo{} or \answerNA{} answer to this question will not be perceived well by the reviewers. 
        \item The claims made should match theoretical and experimental results, and reflect how much the results can be expected to generalize to other settings. 
        \item It is fine to include aspirational goals as motivation as long as it is clear that these goals are not attained by the paper. 
    \end{itemize}

\item {\bf Limitations}
    \item[] Question: Does the paper discuss the limitations of the work performed by the authors?
    \item[] Answer: \answerYes{} 
    \item[] Justification: We conclude our submission with a discussion on some shortcomings of our library with regards to usability, algorithmic benchmarking, and optimization.
    \item[] Guidelines:
    \begin{itemize}
        \item The answer \answerNA{} means that the paper has no limitation while the answer \answerNo{} means that the paper has limitations, but those are not discussed in the paper. 
        \item The authors are encouraged to create a separate ``Limitations'' section in their paper.
        \item The paper should point out any strong assumptions and how robust the results are to violations of these assumptions (e.g., independence assumptions, noiseless settings, model well-specification, asymptotic approximations only holding locally). The authors should reflect on how these assumptions might be violated in practice and what the implications would be.
        \item The authors should reflect on the scope of the claims made, e.g., if the approach was only tested on a few datasets or with a few runs. In general, empirical results often depend on implicit assumptions, which should be articulated.
        \item The authors should reflect on the factors that influence the performance of the approach. For example, a facial recognition algorithm may perform poorly when image resolution is low or images are taken in low lighting. Or a speech-to-text system might not be used reliably to provide closed captions for online lectures because it fails to handle technical jargon.
        \item The authors should discuss the computational efficiency of the proposed algorithms and how they scale with dataset size.
        \item If applicable, the authors should discuss possible limitations of their approach to address problems of privacy and fairness.
        \item While the authors might fear that complete honesty about limitations might be used by reviewers as grounds for rejection, a worse outcome might be that reviewers discover limitations that aren't acknowledged in the paper. The authors should use their best judgment and recognize that individual actions in favor of transparency play an important role in developing norms that preserve the integrity of the community. Reviewers will be specifically instructed to not penalize honesty concerning limitations.
    \end{itemize}

\item {\bf Theory assumptions and proofs}
    \item[] Question: For each theoretical result, does the paper provide the full set of assumptions and a complete (and correct) proof?
    \item[] Answer: \answerNA{} 
    \item[] Justification: Our work does not include novel theoretical results.
    \item[] Guidelines:
    \begin{itemize}
        \item The answer \answerNA{} means that the paper does not include theoretical results. 
        \item All the theorems, formulas, and proofs in the paper should be numbered and cross-referenced.
        \item All assumptions should be clearly stated or referenced in the statement of any theorems.
        \item The proofs can either appear in the main paper or the supplemental material, but if they appear in the supplemental material, the authors are encouraged to provide a short proof sketch to provide intuition. 
        \item Inversely, any informal proof provided in the core of the paper should be complemented by formal proofs provided in appendix or supplemental material.
        \item Theorems and Lemmas that the proof relies upon should be properly referenced. 
    \end{itemize}

    \item {\bf Experimental result reproducibility}
    \item[] Question: Does the paper fully disclose all the information needed to reproduce the main experimental results of the paper to the extent that it affects the main claims and/or conclusions of the paper (regardless of whether the code and data are provided or not)?
    \item[] Answer: \answerYes{}{} 
    \item[] Justification: Appendix~\ref{supp:baselines} of our work includes the hyperparameters and training settings required to replicate the experimental algorithmic performance findings in our paper. Our benchmarking performance results are replicable provided a user has access to the hardware described in Section~\ref{sec:benchmarking}.
    \item[] Guidelines:
    \begin{itemize}
        \item The answer \answerNA{} means that the paper does not include experiments.
        \item If the paper includes experiments, a \answerNo{} answer to this question will not be perceived well by the reviewers: Making the paper reproducible is important, regardless of whether the code and data are provided or not.
        \item If the contribution is a dataset and\slash or model, the authors should describe the steps taken to make their results reproducible or verifiable. 
        \item Depending on the contribution, reproducibility can be accomplished in various ways. For example, if the contribution is a novel architecture, describing the architecture fully might suffice, or if the contribution is a specific model and empirical evaluation, it may be necessary to either make it possible for others to replicate the model with the same dataset, or provide access to the model. In general. releasing code and data is often one good way to accomplish this, but reproducibility can also be provided via detailed instructions for how to replicate the results, access to a hosted model (e.g., in the case of a large language model), releasing of a model checkpoint, or other means that are appropriate to the research performed.
        \item While NeurIPS does not require releasing code, the conference does require all submissions to provide some reasonable avenue for reproducibility, which may depend on the nature of the contribution. For example
        \begin{enumerate}
            \item If the contribution is primarily a new algorithm, the paper should make it clear how to reproduce that algorithm.
            \item If the contribution is primarily a new model architecture, the paper should describe the architecture clearly and fully.
            \item If the contribution is a new model (e.g., a large language model), then there should either be a way to access this model for reproducing the results or a way to reproduce the model (e.g., with an open-source dataset or instructions for how to construct the dataset).
            \item We recognize that reproducibility may be tricky in some cases, in which case authors are welcome to describe the particular way they provide for reproducibility. In the case of closed-source models, it may be that access to the model is limited in some way (e.g., to registered users), but it should be possible for other researchers to have some path to reproducing or verifying the results.
        \end{enumerate}
    \end{itemize}

\item {\bf Open access to data and code}
    \item[] Question: Does the paper provide open access to the data and code, with sufficient instructions to faithfully reproduce the main experimental results, as described in supplemental material?
    \item[] Answer: \answerYes{} 
    \item[] Justification: Our submission includes a link to a public anonymous Github (\url{https://github.com/bluffjax/bluffjax}) which contains didactic examples and tutorials. We also submit our code as supplementary material. 
    \item[] Guidelines:
    \begin{itemize}
        \item The answer \answerNA{} means that paper does not include experiments requiring code.
        \item Please see the NeurIPS code and data submission guidelines (\url{https://neurips.cc/public/guides/CodeSubmissionPolicy}) for more details.
        \item While we encourage the release of code and data, we understand that this might not be possible, so \answerNo{} is an acceptable answer. Papers cannot be rejected simply for not including code, unless this is central to the contribution (e.g., for a new open-source benchmark).
        \item The instructions should contain the exact command and environment needed to run to reproduce the results. See the NeurIPS code and data submission guidelines (\url{https://neurips.cc/public/guides/CodeSubmissionPolicy}) for more details.
        \item The authors should provide instructions on data access and preparation, including how to access the raw data, preprocessed data, intermediate data, and generated data, etc.
        \item The authors should provide scripts to reproduce all experimental results for the new proposed method and baselines. If only a subset of experiments are reproducible, they should state which ones are omitted from the script and why.
        \item At submission time, to preserve anonymity, the authors should release anonymized versions (if applicable).
        \item Providing as much information as possible in supplemental material (appended to the paper) is recommended, but including URLs to data and code is permitted.
    \end{itemize}

\item {\bf Experimental setting/details}
    \item[] Question: Does the paper specify all the training and test details (e.g., data splits, hyperparameters, how they were chosen, type of optimizer) necessary to understand the results?
    \item[] Answer: \answerYes{} 
    \item[] Justification: Our training and test details are shared in Appendix~\ref{supp:baselines} and within the attached code.
    \item[] Guidelines:
    \begin{itemize}
        \item The answer \answerNA{} means that the paper does not include experiments.
        \item The experimental setting should be presented in the core of the paper to a level of detail that is necessary to appreciate the results and make sense of them.
        \item The full details can be provided either with the code, in appendix, or as supplemental material.
    \end{itemize}

\item {\bf Experiment statistical significance}
    \item[] Question: Does the paper report error bars suitably and correctly defined or other appropriate information about the statistical significance of the experiments?
    \item[] Answer: \answerYes{} 
    \item[] Justification: Our performance and algorithmic benchmark results include distributional information as standard errors over multiple seeds. Note that these bounds are not visible on the scale of performance benchmarking.
    \item[] Guidelines:
    \begin{itemize}
        \item The answer \answerNA{} means that the paper does not include experiments.
        \item The authors should answer \answerYes{} if the results are accompanied by error bars, confidence intervals, or statistical significance tests, at least for the experiments that support the main claims of the paper.
        \item The factors of variability that the error bars are capturing should be clearly stated (for example, train/test split, initialization, random drawing of some parameter, or overall run with given experimental conditions).
        \item The method for calculating the error bars should be explained (closed form formula, call to a library function, bootstrap, etc.)
        \item The assumptions made should be given (e.g., Normally distributed errors).
        \item It should be clear whether the error bar is the standard deviation or the standard error of the mean.
        \item It is OK to report 1-sigma error bars, but one should state it. The authors should preferably report a 2-sigma error bar than state that they have a 96\% CI, if the hypothesis of Normality of errors is not verified.
        \item For asymmetric distributions, the authors should be careful not to show in tables or figures symmetric error bars that would yield results that are out of range (e.g., negative error rates).
        \item If error bars are reported in tables or plots, the authors should explain in the text how they were calculated and reference the corresponding figures or tables in the text.
    \end{itemize}

\item {\bf Experiments compute resources}
    \item[] Question: For each experiment, does the paper provide sufficient information on the computer resources (type of compute workers, memory, time of execution) needed to reproduce the experiments?
    \item[] Answer: \answerYes{} 
    \item[] Justification: The exact GPU and CPU setups and memory requirements are discussed in the paper.
    \item[] Guidelines:
    \begin{itemize}
        \item The answer \answerNA{} means that the paper does not include experiments.
        \item The paper should indicate the type of compute workers CPU or GPU, internal cluster, or cloud provider, including relevant memory and storage.
        \item The paper should provide the amount of compute required for each of the individual experimental runs as well as estimate the total compute. 
        \item The paper should disclose whether the full research project required more compute than the experiments reported in the paper (e.g., preliminary or failed experiments that didn't make it into the paper). 
    \end{itemize}
    
\item {\bf Code of ethics}
    \item[] Question: Does the research conducted in the paper conform, in every respect, with the NeurIPS Code of Ethics \url{https://neurips.cc/public/EthicsGuidelines}?
    \item[] Answer: \answerYes{} 
    \item[] Justification: Our work conforms to the Code of Ethics provided. We should note that as a benchmarking tool centered around games with hidden information, our library may be used for the training of AI agents with the explicit goal of deception. We hereby state that BluffJAX is solely intended as a research tool and should not be used for enabling deceptive behavior in the real world or for-profit applications.
    \item[] Guidelines:
    \begin{itemize}
        \item The answer \answerNA{} means that the authors have not reviewed the NeurIPS Code of Ethics.
        \item If the authors answer \answerNo, they should explain the special circumstances that require a deviation from the Code of Ethics.
        \item The authors should make sure to preserve anonymity (e.g., if there is a special consideration due to laws or regulations in their jurisdiction).
    \end{itemize}

\item {\bf Broader impacts}
    \item[] Question: Does the paper discuss both potential positive societal impacts and negative societal impacts of the work performed?
    \item[] Answer: \answerYes{} 
    \item[] Justification: The previous checklist point discusses potential misuse of our library which may result in negative societal impact.
    \item[] Guidelines:
    \begin{itemize}
        \item The answer \answerNA{} means that there is no societal impact of the work performed.
        \item If the authors answer \answerNA{} or \answerNo, they should explain why their work has no societal impact or why the paper does not address societal impact.
        \item Examples of negative societal impacts include potential malicious or unintended uses (e.g., disinformation, generating fake profiles, surveillance), fairness considerations (e.g., deployment of technologies that could make decisions that unfairly impact specific groups), privacy considerations, and security considerations.
        \item The conference expects that many papers will be foundational research and not tied to particular applications, let alone deployments. However, if there is a direct path to any negative applications, the authors should point it out. For example, it is legitimate to point out that an improvement in the quality of generative models could be used to generate Deepfakes for disinformation. On the other hand, it is not needed to point out that a generic algorithm for optimizing neural networks could enable people to train models that generate Deepfakes faster.
        \item The authors should consider possible harms that could arise when the technology is being used as intended and functioning correctly, harms that could arise when the technology is being used as intended but gives incorrect results, and harms following from (intentional or unintentional) misuse of the technology.
        \item If there are negative societal impacts, the authors could also discuss possible mitigation strategies (e.g., gated release of models, providing defenses in addition to attacks, mechanisms for monitoring misuse, mechanisms to monitor how a system learns from feedback over time, improving the efficiency and accessibility of ML).
    \end{itemize}
    
\item {\bf Safeguards}
    \item[] Question: Does the paper describe safeguards that have been put in place for responsible release of data or models that have a high risk for misuse (e.g., pre-trained language models, image generators, or scraped datasets)?
    \item[] Answer: \answerYes{} 
    \item[] Justification: In its current state, BluffJAX is primarily suitable for research purposes and is designed to not be useful for any potential harmful applications as discussed previously.
    \item[] Guidelines:
    \begin{itemize}
        \item The answer \answerNA{} means that the paper poses no such risks.
        \item Released models that have a high risk for misuse or dual-use should be released with necessary safeguards to allow for controlled use of the model, for example by requiring that users adhere to usage guidelines or restrictions to access the model or implementing safety filters. 
        \item Datasets that have been scraped from the Internet could pose safety risks. The authors should describe how they avoided releasing unsafe images.
        \item We recognize that providing effective safeguards is challenging, and many papers do not require this, but we encourage authors to take this into account and make a best faith effort.
    \end{itemize}

\item {\bf Licenses for existing assets}
    \item[] Question: Are the creators or original owners of assets (e.g., code, data, models), used in the paper, properly credited and are the license and terms of use explicitly mentioned and properly respected?
    \item[] Answer: \answerYes{} 
    \item[] Justification: BluffJAX is an original, open-source software library under the MIT license. No external code, data, or models were used in the creation of this work.
    \item[] Guidelines:
    \begin{itemize}
        \item The answer \answerNA{} means that the paper does not use existing assets.
        \item The authors should cite the original paper that produced the code package or dataset.
        \item The authors should state which version of the asset is used and, if possible, include a URL.
        \item The name of the license (e.g., CC-BY 4.0) should be included for each asset.
        \item For scraped data from a particular source (e.g., website), the copyright and terms of service of that source should be provided.
        \item If assets are released, the license, copyright information, and terms of use in the package should be provided. For popular datasets, \url{paperswithcode.com/datasets} has curated licenses for some datasets. Their licensing guide can help determine the license of a dataset.
        \item For existing datasets that are re-packaged, both the original license and the license of the derived asset (if it has changed) should be provided.
        \item If this information is not available online, the authors are encouraged to reach out to the asset's creators.
    \end{itemize}

\item {\bf New assets}
    \item[] Question: Are new assets introduced in the paper well documented and is the documentation provided alongside the assets?
    \item[] Answer: \answerYes{} 
    \item[] Justification: Our open-source library is provided with appropriate documentation for ease of training, benchmarking, and extensibility. 
    \item[] Guidelines:
    \begin{itemize}
        \item The answer \answerNA{} means that the paper does not release new assets.
        \item Researchers should communicate the details of the dataset\slash code\slash model as part of their submissions via structured templates. This includes details about training, license, limitations, etc. 
        \item The paper should discuss whether and how consent was obtained from people whose asset is used.
        \item At submission time, remember to anonymize your assets (if applicable). You can either create an anonymized URL or include an anonymized zip file.
    \end{itemize}

\item {\bf Crowdsourcing and research with human subjects}
    \item[] Question: For crowdsourcing experiments and research with human subjects, does the paper include the full text of instructions given to participants and screenshots, if applicable, as well as details about compensation (if any)? 
    \item[] Answer: \answerNA{} 
    \item[] Justification: Our work does not involve crowdsourcing nor research with human subjects.
    \item[] Guidelines:
    \begin{itemize}
        \item The answer \answerNA{} means that the paper does not involve crowdsourcing nor research with human subjects.
        \item Including this information in the supplemental material is fine, but if the main contribution of the paper involves human subjects, then as much detail as possible should be included in the main paper. 
        \item According to the NeurIPS Code of Ethics, workers involved in data collection, curation, or other labor should be paid at least the minimum wage in the country of the data collector. 
    \end{itemize}

\item {\bf Institutional review board (IRB) approvals or equivalent for research with human subjects}
    \item[] Question: Does the paper describe potential risks incurred by study participants, whether such risks were disclosed to the subjects, and whether Institutional Review Board (IRB) approvals (or an equivalent approval/review based on the requirements of your country or institution) were obtained?
    \item[] Answer: \answerNA{} 
    \item[] Justification: Our work does not involve crowdsourcing nor research with human subjects.
    \item[] Guidelines:
    \begin{itemize}
        \item The answer \answerNA{} means that the paper does not involve crowdsourcing nor research with human subjects.
        \item Depending on the country in which research is conducted, IRB approval (or equivalent) may be required for any human subjects research. If you obtained IRB approval, you should clearly state this in the paper. 
        \item We recognize that the procedures for this may vary significantly between institutions and locations, and we expect authors to adhere to the NeurIPS Code of Ethics and the guidelines for their institution. 
        \item For initial submissions, do not include any information that would break anonymity (if applicable), such as the institution conducting the review.
    \end{itemize}

\item {\bf Declaration of LLM usage}
    \item[] Question: Does the paper describe the usage of LLMs if it is an important, original, or non-standard component of the core methods in this research? Note that if the LLM is used only for writing, editing, or formatting purposes and does \emph{not} impact the core methodology, scientific rigor, or originality of the research, declaration is not required.
    \item[] Answer: \answerNA{} 
    \item[] Justification: The core development of this library does not involve LLMs for any novel, important, or non-standard components.
    \item[] Guidelines:
    \begin{itemize}
        \item The answer \answerNA{} means that the core method development in this research does not involve LLMs as any important, original, or non-standard components.
        \item Please refer to our LLM policy in the NeurIPS handbook for what should or should not be described.
    \end{itemize}

\end{enumerate}